\documentclass[journal]{IEEEtran}
\usepackage{cite}
\usepackage{amssymb,amsfonts}
\usepackage{textcomp}
\usepackage{amsmath,amsthm,graphicx}

\usepackage{url}
\usepackage{xcolor}
\definecolor{green}{HTML}{00B090}
\definecolor{pink}{HTML}{F00090}
\definecolor{purple}{HTML}{6030B0}
\definecolor{black}{HTML}{000000}
\definecolor{palepurple}{HTML}{7864B5}
\definecolor{lilac}{HTML}{D8C4FF}
\definecolor{rose}{HTML}{FEDEFF}
\definecolor{amber}{HTML}{EFBF40}
\definecolor{black}{HTML}{000000}
\definecolor{white}{HTML}{FFFFFF}
\usepackage{hyperref}
\hypersetup{
    colorlinks=true,
    citecolor=pink,
    linkcolor=purple,
    filecolor=palepurple,      
    urlcolor=green
    }
\usepackage{colortbl}
\newcolumntype{P}{>{\columncolor{palepurple}}c}
\newcolumntype{R}{>{\columncolor{rose}}c}
\newcolumntype{L}{>{\columncolor{lilac}}c}
\usepackage{multirow}
\usepackage{enumitem}
\usepackage{titlesec}
\usepackage{wrapfig}				
\usepackage{bm}
\usepackage{cancel}
\usepackage{caption}
\usepackage{subcaption}
\usepackage{multirow}
\usepackage{booktabs}

\usepackage{algorithm}
\usepackage[noend]{algpseudocode}
\makeatletter

\algrenewcommand\algorithmicrequire{\textbf{Input:}}
\algrenewcommand\algorithmicensure{\textbf{Output:}}

\newcommand{\cm}[2]{
\begin{equation}
\begin{aligned}
#1
\end{aligned}
#2
\end{equation}
}

\begin{document}
\title{
Integrated Gaussian Processes for Robust and Adaptive Multi-Object Tracking
}
\author{Fred~Lydeard$^\dagger$,~Bashar~I.~Ahmad$^\dagger$,~and~Simon~Godsill$^\dagger$,~\IEEEmembership{Fellow,~IEEE}\\
\textit{$^\dagger$Probabilistic Systems, Information, and Inference Group, Department of Engineering,\\ University of Cambridge, Cambridge, UK\\\vspace{-5mm}}%
\thanks{Manuscript received XXXXX XX, 2025; revised XXXX XX, 2025.}}
\markboth{IEEE Transactions on ...,~Vol.~0, No.~0, XXXXXXX~2025}%
{Lydeard \MakeLowercase{\textit{et al.}}: Integrated Gaussian Processes for Robust and Adaptive Multi-Object Tracking}
\maketitle
\begin{abstract}
This paper presents a computationally efficient multi-object tracking approach that can minimise track breaks (e.g., in challenging environments and against agile targets), learn the measurement model parameters on-line (e.g., in dynamically changing scenes) and infer the class of the tracked objects, \textit{if} joint tracking and kinematic behaviour classification is sought. It capitalises on the flexibilities offered by the integrated Gaussian process as a motion model and the convenient statistical properties of non-homogeneous Poisson processes as a suitable observation model. This can be combined with the proposed effective track revival / stitching mechanism. We accordingly introduce the two robust and adaptive trackers, Gaussian and Poisson Process with Classification (GaPP-Class) and GaPP with Revival and Classification (GaPP-ReaCtion). They employ an appropriate particle filtering inference scheme that efficiently integrates track management and hyperparameter learning (including the object class, if relevant). GaPP-ReaCtion extends GaPP-Class with the addition of a Markov Chain Monte Carlo kernel applied to each particle permitting track revival and stitching (e.g., within a few time steps after deleting a trajectory). Performance evaluation and benchmarking using synthetic and real data show that GaPP-Class and GaPP-ReaCtion outperform other state-of-the-art tracking algorithms. For example, GaPP-ReaCtion significantly reduces track breaks (e.g., by around 30\% from real radar data and markedly more from simulated data). 

\end{abstract}

\section{Introduction}
\label{sec:intro}

Multi-object tracking -- inferring dynamically changing quantities (e.g. position, velocity, orientation, or any other spatio-temporal characteristic) of one or more objects -- is a task required in many applications. These include robot navigation, 
group leader classification \cite{li2021sequential}, for example in an animal pack, 
traffic monitoring \cite{garcia2024traffic}, 
to name a few. This is a well-established research area with numerous algorithms, some of which are applied in this paper for performance benchmarking.

Whilst (semi-)autonomous systems, such as drones, are often slow moving and follow optimised trajectories, they can be highly agile and manoeuvrable. This requires the multi-object tracker to handle a wide range of `kinematic' behaviours. Such a capability is often restricted by the tracker using dynamical models with fixed, fine-tuned, parameters, for example, the process noise variance, which describes the level of variability in the object’s motion. Thus, classification of these objects can be essential, and is better facilitated when considered jointly with the tracking \cite{harman2021}. It is often based upon intermittent salient features, such as observing a micro-Doppler effect \cite{ahmad2024review}. Robust tracking which minimises the number track breaks (i.e., changes in track ID) is essential to improve this classification accuracy, since a break may result in a misclassification until an aforementioned salient feature is again available. This strongly motivates a more robust tracker. The ability to retrospectively revive tracks after a short time would help to facilitate this, which is a key contribution of this proposed methodology. 


Another way to reduce track breaks is to use a sufficiently flexible dynamical model. Gaussian Processes (GPs) have many uses including integral estimation in Bayesian numerics \cite{xi2018}
as well as in a variety of object trackers, to infer its location or extent \cite{wahlstrom2015}. The closed form tractability coupled with their flexibility has made GPs appealing in these areas and is what makes them a strong candidate for this tracker focusing on robustness. For usability in real-world scenarios, it is important for the tracker to be adaptive, learning its (dynamic or static) parameters in real time. 

\subsection{Contributions}
\label{ss:contributions}

The contributions of this paper are the establishment of two adaptive and robust multi-object trackers, `GaPP-Class' (\textbf{G}aussian \textbf{a}nd \textbf{P}oisson \textbf{P}rocess with \textbf{Class}ification) and `GaPP-ReaCtion' (GaPP with \textbf{Re}vival \textbf{a}nd \textbf{C}lassifica\textbf{tion}), together with experiments benchmarking their performance. GaPP-Class: 
\begin{itemize}
    \item Uses an integrated GP (iGP) dynamical model for representing a wide range of motion behaviours, and a Non-Homogeneous Poisson Process (NHPP) measurement model, building on the preliminary work in \cite{goodyer2023gapp};
    \item Integrates track management directly into the Bayesian scheme
    \item Enables joint target tracking and classification \textit{if desired}, based upon objects' kinematic behaviour through learning their iGP hyperparameters; and
    \item Presents an improved online learning approach of the observation model hyperparameters which facilitates adapting to dynamically varying environments and sensor characteristics. 
\end{itemize}
GaPP-ReaCtion extends this to further improve robustness through a novel Markov Chain Monte Carlo (MCMC)-based revival process that allows reconnection of broken trajectories, thereby minimising erroneous track breaks. It should be noted that, if classification is not sought, then either of these methods can be used with one class, in which case the associated parameters of the iGP are assumed known.

\subsection{Paper Layout}
\label{ss:paperlayout}
Section \ref{sec:relatedwork} discusses relevant related work, whilst Section \ref{sec:models} outlines the generative model of the scenario. Section \ref{sec:inference} details how to make relevant inferences with GaPP-Class, which is done entirely excluding the `revival' concept of Section \ref{sec:revival} -- an appendage to GaPP-Class to further improve inference, yielding GaPP-ReaCtion. Section \ref{sec:exp} shows experimental results of these methods\footnote{Code available at \url{https://github.com/afredgcam/GaPP-ReaCtion}.}, on both synthetic and real data sets, which is concluded in Section \ref{sec:conclusion}. The appendices contain results used to explain the methodology.

\section{Related Work}
\label{sec:relatedwork}

In Bayesian filtering, an object's state (i.e., its latent quantities of interest) at time step $k$, $\bm x_k$, and any corresponding measurements, $\bm y_k$, are subject to Markovian assumptions (see e.g., \cite{sarkka2013}), leading to the common predict and update steps. Analytic solutions to these filtering equations under linear Gaussian assumptions over both the states and measurements yield the Kalman Filter (KF) \cite{kalman1960new}. This and its non-linear / non-Gaussian extensions, the Extended and Unscented KFs, are single-object methods which are (further) extended for inference over multiple targets and clutter, often by heuristics. 

Typical multi-object trackers include the Multiple Hypothesis Tracker (MHT) \cite{reid1979algorithm}, the Joint Probabilistic Data Association (JPDA) filter \cite{fortmann1980multi} and the Probability Hypothesis Density (PHD) filter \cite{mahler2000}. The first two (and common implementations of the last) impose the assumption that only (at most) one measurement originates from each object. Several more recent trackers continue to make this same assumption, such as the Trajectory Poisson Multi-Bernoulli Mixture (TPMBM) model \cite{granstrom2018poisson} and the time-varying autoregressive tracker of \cite{mcdougall2025target}. The methodology presented here focuses on scenarios in which this is not the case, such as the aforementioned radar-based surveillance.


Alternative tracking frameworks include Message Passing (MP) based methods \cite{meyer2018message}, specifically using a loopy belief propagation algorithm to perform inference. This, along with the GM-PHD, will benchmark the proposed methods. Whilst the JPDA uses a NHPP \cite{gilholm2005} for the clutter, the Rao-Blackwellised Association-Based NHPP (RB-AbNHPP) \cite{li2023adaptive} uses it for all objects as well. Inference is done via Sequential Markov Chain Monte Carlo (SMCMC)
including learning the objects' extents together with their states and all of the (time-varying) parameters associated with the NHPP, except for the measurement error variance, similarly to \cite{li2022}. The proposed work additionally learns this error variance, using a Particle Filter (PF).
Furthermore, the RB-AbNHPP does not innately handle track initialisation and deletion. Whilst a Dirichlet Process (DP) scheme, (e.g., \cite{carevic2016}) has fewer parameters to learn than a NHPP, the lack of interpretability makes choosing priors for them, and hence learning, more difficult. It also would impede the proposed track revival mechanism (more below). The DP does allow for very natural object detection and initialisation, so our use of the NHPP will be accompanied by careful consideration of this.

For simply manoeuvring objects, the (nearly) Constant Velocity (CV) model is often sufficient. More manoeuvrability is provided by the Constant Acceleration (CA) model or variants like the Singer model .
The Interacting Multiple Model (IMM) 
acts as a kinematic switching model, allowing an object to exhibit many behaviours. The choice of models is non-trivial, and the number of parameters can be prohibitively vast. 
More flexible dynamics are permitted by L\'evy process-based models \cite{gan2021levy, kindap2023generalised}, but these can be prone to extremely heavy-tailed `teleporting' behaviour. 

To achieve the desired flexibility whilst minimising user-input, GPs can be used for the dynamics. A SE GP has been used to directly model the location of an object in \cite{aftab2020}, but the stationarity of the SE kernel is generally unsuitable \cite{lydeard2025integrated}. Applying a GP to the disturbances -- the differences in an object's state between consecutive time steps -- is more realistic as an object's velocity is likely approximately stationary \cite{goodyer2023gapp, goodyer2023flexible, sun2022gaussian}.

The iGP, specifically with the integrated SE (iSE) kernel \cite{lydeard2025integrated}, provides a simple yet suitable GP dynamical model for our proposed tracker, the parameters of which are discretely learnt. Due to the inherent flexibility of the iSE model it is conjectured here that, in some circumstances, the iSE parameters are properties of an object's class, thus, the availability of these discrete options for the iSE parameters is feasible. In these scenarios, this tracker would be jointly tracking and classifying the objects based on their kinematic behaviour. 
Many methods define classes by different (parametrised) switching multiple models (a precursor to the IMM), differing by model choice \cite{boers2001integrated,maskell2004joint} or model parameters \cite{gordon2002efficient,zhang2024motion}. 
In extended tracking, the extent can be used to discriminate between classes \cite{wang2024joint,cheng2024multiple}, or other features \cite{he2014joint,angelova2006joint}.
A `class likelihood' permits additional weighting of the class probabilities \cite{xue2023novel,li2022loopy}. 

As mentioned with the IMM, choice of models and their parameters in multiple models is difficult. This is exacerbated when doing this generally for each object class, as they should capture the full range of motion of that object type, requiring a large number of models in the mixture. The flexibility of the (i)GP models used here allow for a single parameter specification to capture the significant majority of the possible manoeuvres of an object class: the iSE method requires just two hyperparameters per class. Whilst learning the hyperparameters of the iGP can be used for classification in some scenarios, such as the drone surveillance example considered, this is merely serendipitous to the tracker being able to learn these hyperparameters in order to be more adaptive.

This work will present a MCMC-based `track revival' mechanism, to explicitly reduce the number of track breaks in near-real-time (i.e., with only a short temporal delay). Track revival, sometimes referred to in the literature as track stitching, has many uses, such as material reconstruction \cite{czabaj2014three}.
Track revival was discussed as a potential benefit to a message passing-based multi-target tracker \cite{chen2009efficient}, but the example shown assumed a known number of objects, significantly simplifying the task. 

A common track revival technique for radar data is to, in post-processing, extend tracks in time (both forwards and backwards) and compatible tracks are stitched \cite{liang2023augmented, li2022graph}. A near-real-time revival scheme was presented in \cite{xu2022move} as part of a Labelled Multi-Bernoulli (LMB) filter, in which measurements which fall within the gate of a terminated track can revive that track. Because the LMB filter assumes at most one observation per object per time, this revival mechanism may be error-prone under higher clutter rates.

\section{Generative Models}
\label{sec:models}

\subsection{Dynamical Model}
\label{ss:dynamicalmodel}

The dynamical model of each object is an approximation of an integrated Gaussian Process (iGP), specifically, the integrated Squared Exponential (iSE) with the second Markovian approximation of size $d'$ seconds \cite{lydeard2025integrated}. As noted there, this model is suitable for irregular time steps, but, as is also done there, we opt for constant time steps of size $T$ seconds, and an initial time of $t^* = 0$, which is at time step $k=0$, for simplicity of illustration. Due to the constant time step size, we can formulate the following in terms of $d = \left\lfloor\frac{d'}{T}\right\rfloor$, the (approximate) number of time steps lasting $d'$ seconds. For further simplicity, we assume 1-dimensional positions in the understanding that, for more dimensions, this can be applied independently to each.

For $X_{ki}:=X_i(kT)$ the 1D position of the $i^\text{th}$ object at time step $k$, then this object's state at time step $k$ is $\bm x_{ki} = (X_{ki},\dots,X_{k-d+1,i})^T$, where it is assumed that the object has been active for at least $d$ time steps. Define $\mathcal N(x\mid a,b^2)$ as the probability density function (PDF) of a $\mathcal N(a,b^2)$ random variable / vector, and let $\Phi(x)$ be the cumulative distribution function (CDF) of the standard Gaussian distribution. Observe that the Squared Exponential (SE) covariance function, with hyperparameters, $\sigma^2$ and $\ell$, can be written as
\cm{
\ddot C(t,t'\mid \sigma^2,\ell) &= \sigma^2\exp\left\{-\frac{(t-t')^2}{2\ell^2}\right\}\\
&= \sqrt{2\pi}\ell\sigma^2\mathcal N(t\mid t',\ell^2).
}{\label{eq:rewriteSE}}
Further, define
\cm{
\Xi_b(x\mid a) &= \int \Phi\left(\frac{x-a}{b}\right)\ dx\\
&= (x-a)\Phi\left(\frac{x-a}{b}\right) + b^2\mathcal N(x\mid a,b^2),
}{\label{eq:intPhi}}
where the last line follows from integration by parts \cite{lydeard2025integrated}. Then, the SE kernel can be twice integrated to get
\cm{
\!\!C&(t,t'\mid\sigma^2,\ell) := \int_0^t\int_0^{t'} \ddot C(\tau,\tau'\mid \sigma^2,\ell)\ d\tau'd\tau\\
&= \sqrt{2\pi}\ell\sigma^2\left(\Xi_\ell(t\mid0) + \Xi_\ell(0\mid t') - \Xi_\ell(t\mid t')\right) - \sigma^2\ell^2.
}{\label{eq:iSEkernel}}
Finally, for arbitrary vectors of inputs, $\bm x$ and $\bm y$, with lengths, $n$ and $m$, respectively, we shall let $f_{\bm x} = (f(x_1),\dots,f(x_n))^T\in\mathbb R^n$ and $g_{\bm x\bm y} = (g_{\bm x,y_1},\dots,g_{\bm x,y_m})\in\mathbb R^{n\times m}$ for any univariate function, $f$, and bivariate function, $g$.

Now, our chosen iSE approximation is expressible as a linear state space model,
\cm{
\bm x_{ki} &= F_{ki}\bm x_{k-1,i} + \bm q_{ki},\\
\bm q_{ki} &\sim \mathcal N(\bm 0_d, Q_{ki}),
}{\label{eq:linearstatespaceiSE1}}
where $\bm 0_\alpha\in\mathbb R^\alpha$ is a vector / matrix of zeroes (similarly for $\bm 1_\alpha$), and 
\cm{
F_{ki} &= 
\begin{pmatrix}
    \bm f_{ki}^T & 1-\bm f_{ki}^T\bm 1_{d-1}\\
    I_{d-1} & \bm 0_{d-1}
\end{pmatrix},\\
Q_{ki} &= q'_{ki}\bm e_{1d}\bm e_{1d}^T,\\
\bm f_{ki} &= C_{\bm t_{d-1}\bm t_{d-1}}^{-1}C_{\bm t_{d-1},dT},\\
q'_{ki} &= C_{dT,dT} - C_{dT,\bm t_{d-1}}\,\bm f_{ki},
}{\label{eq:linearstatespaceiSE2}}
with $I_\alpha\in\mathbb R^{\alpha\times\alpha}$ being the identity matrix, $\bm e_{ab}\in\mathbb R^b$ being the dimension $a\le b$ unit vector (that is, $(\bm e_{ab})_i = 1$ if and only if $i=a$ and is 0 otherwise) and $C$ being evaluated with $\sigma_{ki}^2$ and $\ell_{ki}$ as the hyperparameters. Here, $\bm t_\alpha = T(\alpha,\dots,1)^T$. For GaPP-ReaCtion, $\sigma_{ki}^2$ and $\ell_{ki}$ will be time independent, which will ultimately make $F_{ki}$ and $Q_{ki}$ time independent, also.

\subsection{Observation Model}
\label{ss:obsmodel}

The observation model is a NHPP \cite{gilholm2005}, wherein the number of observations of each of the $N_k$ objects, or of clutter (henceforth indexed by $i=0$) is independently drawn from a Poisson distribution, and then the value of each of these is drawn from some likelihood distribution. We choose this likelihood as Gaussian for objects, but, for clutter, leave it unspecified. That is, for all $i \in \{0,\dots,N_k\}$
\cm{
n_{ki}\sim \text{Poisson}(\mu_{ki}),
}{\label{eq:nhpp1}}
and, defining $a_{kl}$ as the association of datum $y_{kl}$, we have 
\cm{
y_{kl}\mid a_{kl}=i,\bm x_{ki}\sim \mathcal N(\bm e_{1d}^T\bm x_{ki},s_k^2),
}{\label{eq:nhpp2}}
for $i\ge 1$, and 
\cm{
y_{kl}\mid a_{kl}=0 \sim p(y),
}{\label{eq:nhpp3}}
for the unspecified clutter distribution, $p(y)$. Also, we will write $n_k = \sum_{i=0}^{N_k} n_{ki}$ as the total number of data. It should be noted, in \eqref{eq:nhpp2}, that $\bm e_{1d}^T\bm x_{ki} = X_{ki}$, the current location of the object, and that $s_k^2$ is the measurement variance at time step $k$.

\subsection{Initialisation and Deletion Models}
\label{ss:initdelmodels}

At each time step, objects can be initialised, or `birthed'. Let $\eta_k$ be the number of such births at step $k$. We assume a NHPP over the new objects. Thus, 
\cm{
\eta_k\sim \text{Poisson}(\gamma_k),
}{\label{eq:birthingdistribution}}
for some birth rate parameter, $\gamma_k$, with the location of each new object being sampled as
\cm{
X_i^*\sim p(X^*).
}{\label{eq:birthdistribution}}
In this work, we will assume a uniform birth distribution over the scene, $S$, so $p(X^*) = \rho\,1(X^*\in S)$, where $\rho^{-1}=\int_SdX$ is the volume of $S$. The birth time step of this object is then $\check\kappa_i = k$.

The dynamical model described in Subsection \ref{ss:dynamicalmodel} applies only when the object has been active for at least $d$ time steps. Until then, a modification is used, whereby the state of the object $\bm x_{ki} = (X_{ki},\dots,X_{\check\kappa_ii})^T$ and follows a similar linear state space model as that described by \eqref{eq:linearstatespaceiSE1} and \eqref{eq:linearstatespaceiSE2}, but where $G_{ki}$ replaces $F_{ki}$, with
\cm{
G_{ki} &= 
\begin{pmatrix}
    \bm g_{ki}^T + (1 - \bm g_{ki}^T\bm 1_{k-\check\kappa_i})\bm e_{k-\check\kappa_i,k-\check\kappa_i}^T\\
    I_{k-\check\kappa_i}
\end{pmatrix},\\
\bm g_{ki} &= C_{\bm t_{k-\check\kappa_i}\bm t_{k-\check\kappa_i}}^{-1}C_{\bm t_{k-\check\kappa_i},(k-\check\kappa_i+1)T},
}{\label{eq:growingstatespace1}}
and, similarly, $\bm r_{ki}$ replaces $\bm q_{ki}$, with
\cm{
\bm r_{ki}&\sim \mathcal N(\bm 0_{k-\check\kappa_i+1},R_{ki}),\\
R_{ki} &= r'_{ki}\bm e_{1,k-\check\kappa_i+1}\bm e_{1,k-\check\kappa_i+1}^T,\\
r'_{ki} &= C_{(k-\check\kappa_i+1)T,(k-\check\kappa_i+1)T} - C_{(k-\check\kappa_i+1)T,\bm t_{k-\check\kappa_i}} \bm g_{ki}.
}{\label{eq:growingstatespace2}}

Objects can become inactive or `get deleted', and once this happens, they cannot ever become active again. It should be noted that the `revival' aspect of GaPP-ReaCtion refers to the ability of any inferred tracks which have been deleted to be revived. The idea, as will be later discussed, is that the track may be accidentally deleted when the true object is still active, and then the algorithm may initialise a second track for this same object, constituting an undesirable track break. 

The activity of object $i$ at time step $k$ is denoted by $\zeta_{ki}\in \mathbb Z_2$. If $i$ was active at time step $k-1$, that is, $\zeta_{k-1,i}=1$, then it remains active with some survival probability $\psi_{ki}$. This can be summarised by
\cm{
\zeta_{ki}\mid \zeta_{k-1,i}\sim \text{Bernoulli}(\zeta_{k-1,i}\psi_{ki}).
}{\label{eq:deletemodel}}
Furthermore, freshly initialised objects are always active, meaning that $\zeta_{\check\kappa_i i}=1$ with certainty, for all $i$.
The time step of deletion for object $i$ is denoted $\kappa_i$, and is the first time step for which $\zeta_{ki}=0$.

\subsection{Object Classes}
\label{ss:classes}

It is assumed here that the object class can be represented by the hyperparameters used in the dynamical model. For example, in surveillance applications, we may assume $\sigma^2 = 30$ and $\ell = 4$ for faster, larger and less agile objects, such as an airliner, but slower and potentially highly manoeuvring objects, such as rotary-wing drones, have $\sigma^2=5$ and $\ell = \frac12$. It should be noted that equivalent assumptions are not available if using a zero-mean GP with an isotropic covariance function, such as the SE in \cite{aftab2020}, because the hyperparameters will be dependent on the locations of the object. Let the class label (out of a nominal discrete set) of the track object $i$ be $\mathcal C_i$. Then, the dynamical model hyperparameters of object $i$ are simply $\sigma_{\mathcal C_i}^2$ and $\ell_{\mathcal C_i}$. The hyperparameters are class dependent and dictate the transition matrices, $F_{ki}$ and $Q_{ki}$ from \eqref{eq:linearstatespaceiSE1} as well as $G_{ki}$ from \eqref{eq:growingstatespace1} and $R_{ki}$ from \eqref{eq:growingstatespace2}. They shall be hereby denoted $F_c, Q_c, G_c$ and $R_c$, respectively, for brevity. The prior probabilities of an object being from each class (i.e. having a set of hyperparameters) is given by the vector $\bm\pi_+$.

\section{Inference}
\label{sec:inference}

Inference based on the models of Section \ref{sec:models} is done here via the use of a particle filter with $J$ particles, over the number of object initialisations at each time step, the activity of the objects, and the data associations. So, the collection of particles at time step $k$ is $\{\eta_{1:k}^j,\bm \zeta_{1:k}^j,\bm a_{1:k}^j\}_{j=1}^J$. It should be noted that this inference easily generalises for use with any (Cartesian) Gaussian dynamical model with the location in (the first position of) the state, and its measurements simply being noisy observations of that location. The only changes to the scheme will be to the predict and update steps.

\subsection{Hyperparameters}
\label{ss:hyperpars}


Most (hyper)parameters are assumed unknown, and are learnt here. Whilst no prior knowledge of the model that governs their evolution over time is assumed available in this paper, we consider slowly changing parameters or those that have an underlying piece-wise constant trend, thus assumed to remain fixed over modest durations of time. This is reasonable in numerous scenarios such as with long range sensors (e.g., radars) where environmental parameters (e.g., clutter characteristics) can change relatively slowly over time compared to the sensor update rate. With that in mind, this work initially just considers fixed parameters, and the small adaptation required to account for gradual parameter dynamicism is explained afterwards. Thus, for now, we consider $\mu_i:=\mu_{ki}$ for all $i\in\mathbb N$, $\gamma:=\gamma_k$ and $s^2:=s_k^2$.

Within a given particle, full Bayesian inference is tractable on $\mu_i$ and $\gamma$ if they are given Generalised Inverse Gaussian (GIG) priors, as in \cite{li2023adaptive} -- in the proposed work, we use the more standard special case of GIG that is the Gamma distribution. At time step $k$, our priors are inherited as their respective posteriors from time step $k-1$. Namely,
\cm{
\gamma&\sim\text{Gamma}(\varepsilon_{k-1},\xi_{k-1}),\\
\mu_i&\sim\text{Gamma}(\alpha_{k-1,i},\beta_{k-1,i}),
}{\label{eq:gammapriors}}
for all $i\in\{0,\dots,N_{k-1}\}$ and for some $\varepsilon_{k-1},\xi_{k-1}$ and $\{\alpha_{k-1,i},\beta_{k-1,i}\}_{i=0}^{N_{k-1}}$. Once associations, $\bm a_k$, and the number of new objects, $\eta_k$, have been sampled, the above can be updated. For $\gamma$, it is possible to see that
\cm{
p(\gamma\mid \eta_k) &\propto p(\eta_k\mid\gamma)\,p(\gamma)\\
&= \text{Poisson}(\eta_k\mid \gamma)\,\text{Gamma}(\gamma\mid \varepsilon_{k-1},\xi_{k-1})\\
&\propto \text{Gamma}(\gamma\mid \varepsilon_{k-1}+\eta_k,\xi_{k-1}+1),
}{\label{eq:updategamma1}}
hence, the updated hyperparameters are 
\cm{
\varepsilon_k = \varepsilon_{k-1} + \eta_k, && \xi_k = \xi_{k-1} + 1.
}{\label{eq:updategamma2}}
Similarly, using the association counts, $n_{ki} := |\{l:a_{kl}=i\}|$, for each object (or clutter), and the identical prior and likelihood, we can update the hyperparameters of each $\mu_i$ immediately as
\cm{
\alpha_{ki} = \alpha_{k-1,i} + n_{ki}, && \beta_{ki} = \beta_{k-1,i} + 1.
}{\label{eq:updategamma3}}
This is for an active track, $i$, with $\zeta_{ki}=1$. For completeness, we define $\alpha_{ki}=\alpha_{k-1,i}$ and similarly for $\beta_{ki}$ in the case where $\zeta_{ki}=0$.

Note that any new tracks will be initialised with some global baselines as their prior hyperparameters at their time step of birth, $\check\kappa_i$. We shall denote these baselines as $\alpha_+$ and $\beta_+$. Similarly, we also require baselines for the clutter and birth rates, $\mu_0$ and $\gamma$, which shall be $\alpha'_+,\beta'_+$ and $\varepsilon_+,\xi_+$, respectively, when initialising the model.

The same tractability is not available for $s^2$, even using its (conditional) conjugate prior, the GIG distribution, because marginalising would eliminate the Gaussianity required for analytically tractable tracking, which is assumed to be more desirable. As such, it is better to think of this, not as an approximation to full Bayesian inference, so much as a scheme which claims that $s^2$ is known, but that this knowledge is acquired in a principled manner. That is, a `prior' at time step $k$ is still specified, again as a standard special case of GIG, this time the Inverse Gamma,
\cm{
s^2\sim\text{InvGamma}(\varphi_{k-1},b_{k-1}),
}{\label{eq:invgammaprior}}
where $\varphi_{k-1}$ and $b_{k-1}$ are again the hyperparameters from the most recent posterior. From this, a point estimate is taken ($\mathbb E(s^2) = \frac{b_{k-1}}{\varphi_{k-1}-1}$). The hyperparameters of $s^2$ are updated as follows. Firstly, with $\text{IG}(x\mid a,b)$ as the PDF of an Inverse Gamma distribution with parameters $a$ and $b$, we have
\cm{
p(s^2&\mid \bm y_k,\bm a_k,\{X_{ki}\}_{i=1}^{N_k})\\
&\propto \prod_{i=1}^{N_k} \! \mathcal N(\bm y_{ki}\mid X_{ki}\bm 1_{n_{ki}},s^2I_{n_{ki}})\,\text{IG}(s^2\mid \varphi_{k-1},b_{k-1})\\
&\propto \text{IG}(s^2\mid \varphi'_k, b'_k),
}{\label{eq:updateinvgamma1}}
with
\cm{
\varphi'_k=\varphi_{k-1}+\frac{\sum_{i=1}^{N_k} n_{ki}}2, && b'_k = b_{k-1} +\! \sum_{i=1}^{N_k} \frac{n_{ki}}2 \overline{(y_{ki}-X_{ki})^2}
}{\label{eq:updateinvgamma1.1}}
where, with a slight abuse of vector and set notation, $\bm y_{ki} = \{y_{kl}\mid a_{kl}=i\}$, and $\overline{(y_{ki}-X_{ki})^2}:= \frac1{n_{ki}}\sum_{y\in \bm y_{ki}}(y-X_{ki})^2$. The update term for $b$ is (half of) the total observed variation, whilst for $\varphi$, it is (half of) the degrees of freedom over which the variation occurred. Since the unknown $X_{ki}$ cannot be tractably marginalised, one can estimate the observed variations by the sample variance and its degrees of freedom instead. Specifically, our estimated update is
\cm{
\varphi_k &= \varphi_{k-1} + \frac12\sum_{\{i:n_{ki}>1\}}(n_{ki}-1),\\
b_k &= b_{k-1} + \frac12\sum_{\{i:n_{ki}>1\}}n_{ki}\overline{(y_{ki} - {\bar y}_{ki})^2}\\
&= b_{k-1} + \frac12\sum_{\{l:a_{kl}>0\}}(y_{kl}-{\bar y}_{ka_{kl}})^2.
}{\label{eq:updateinvgamma2}}
The baselines for these are defined as $\varphi_+$ and $b_+$.

The aforementioned changes for learning time-\textit{dependent} parameters concern re-establishing the priors. For fixed parameters, we just take the posteriors from the previous time step as the current priors. Taking $\varepsilon_k$ and $\xi_k$ as illustrative examples, they can be explicitly written as 
\cm{
\varepsilon_k = \varepsilon_+ + \sum_{\kappa=1}^k \eta_\kappa, && \xi_k = \xi_+ + k.
}{\label{eq:explicitgamma1}}
However, for learning a dynamic $\gamma_k$, use of a `forgetting factor', $\lambda_\gamma\in(0,1)$, can be employed, such that the influence of observations decays exponentially. Namely, \eqref{eq:explicitgamma1} becomes
\cm{
\varepsilon_k = \varepsilon_+ + \sum_{\kappa=1}^k \lambda_\gamma^{k-\kappa}\eta_\kappa, && \xi_k = \xi_+ + \frac{1-\lambda_\gamma^k}{1 - \lambda_\gamma}.
}{\label{eq:explicitgamma2}}
Then, with the prior at time step $k+1$ as $\check \varepsilon_{k+1}$, it can be shown that
\cm{
\check\varepsilon_{k+1} = (1 - \lambda_\gamma)\varepsilon_+ + \lambda_\gamma\varepsilon_k,
}{\label{eq:explicitgamma3}}
a weighted average between the previous value and the initial prior, where $\lambda_\gamma$ is the weight towards the previous value. Thus, if $\gamma_k$ is expected to vary gradually, $\lambda_\gamma\approx 1$ is sensible, but if it is changing rapidly, then $\lambda_\gamma\approx 0$ is better. A similar result exists for $\check\xi_{k+1}$. The other parameters, $\mu_{ki}$ for all $i\ge0$ and $s_k^2$, can undergo the same treatment with their own associated forgetting factors. This would be done within the `predict' step, alongside the track predictions in Subsection \ref{ss:trackprediction}. For simplicity, however, the remainder of this piece just assumes static parameters.

It should be noted that the $\psi_i:=\psi_{ki}$ are assumed to be known, since, without any connection between $\psi_i$ and $\psi_\iota$ across objects, then each $\psi_i$ is the parameter of a Geometric distribution of which we get only one full observation (the lifespan of object $i$) and it only comes when that object is deleted, so it can no longer help with the future tracking. If there is an explicit desire to learn it, then a small extension to this model will be needed. Here, however, we simply take a constant $\psi$ for all $k$ and $i$.

Finally, the dynamical model hyperparameters are assumed known for each class, but the class of each object is not known. To learn this, each track has a probability vector, $\bm \pi_{ki}$, corresponding to the classes, $\mathcal C_1,\dots,\mathcal C_{N_c}$. As before, the prior at time step $k$ is $\bm \pi_{k-1,i}$. After the predict step (see Subsection \ref{ss:trackprediction}), each track has a predicted state mean and variance for each class, namely $\check{\bm m}_{kic}$ and $\check V_{kic}$, respectively. Once associations have been sampled, the updated class probabilities can be computed as
\cm{
\mathbb P(\mathcal C_i=c\mid \bm y_{ki}) &\propto p(\bm y_{ki}\mid \mathcal C_i=c)\,\mathbb P(\mathcal C_i=c)\\
\Rightarrow \pi_{kic} &\propto p(\bm y_{ki}\mid \sigma_c^2,\ell_c)\,\pi_{k-1,i,c},
}{\label{eq:updatedpi1}}
where $p(\bm y_{ki}\mid \sigma_c^2,\ell_c) = \mathcal N(\bm y_{ki}\mid \bm \mu_{kic}^y,\Sigma_{kic}^y)$, with
\cm{
\bm\mu_{kic}^y &= (\check{\bm m}_{kic})_1\bm 1_{n_{ki}}\\
\Sigma_{kic}^y &= (\check V_{kic})_{1,1}\bm 1_{n_{ki}\times n_{ki}}+ \frac{b_k}{\varphi_k-1}I_{n_{ki}}.
}{\label{eq:updatedpi2}}
This is only facilitated by our assumption of `knowing' $s^2$. Objects will be initialised with a baseline class probability vector, $\bm\pi_+$, which is the prior over the classes in the generative model. It should be noted that updating the class probabilities for each object individually is justified since the classes of different objects are conditionally independent given the associations, which will have been sampled.

\subsection{Track Deletion}
\label{ss:trackdel}

Track deletion is a two-part procedure. Firstly, the initial survival, $\check\zeta_{ki}$, of each active track, $i$, is a set of heuristic conditions: Here, we choose that if a track with $\zeta_{k-1,i}=1$ (significantly) leaves the field of view, has too high a variance on its current location, or has no associations for sufficiently many consecutive time steps, then it is automatically deleted, leading to $\check\zeta_{ki}=0$. Otherwise, $\check\zeta_{ki} = \zeta_{k-1,i}$. The second stage, leading to the final survival at step $k$, has $\zeta_{ki}=1$ being independently sampled with probability $\psi\check\zeta_{ki}$. 

\subsection{Track Prediction}
\label{ss:trackprediction}

The predict step for track $i$ at time step $k$ consists of computing $\check{\bm m}_{kic}$ and $\check V_{kic}$ for all class indices $c\in\{1,\dots,N_c\}$. Because this is a Gaussian linear state space model, the predict step parameters are available in closed form using \eqref{eq:linearstatespaceiSE2} from the standard Kalman predict step \cite{kalman1960new,sarkka2013},
\cm{
\check{\bm m}_{kic} &= F_c\bm m_{k-1,i,c},\\
\check V_{kic} &= F_c V_{k-1,i,c} F_c^T + Q_c,
}{\label{eq:trackpred1}}
or, if $\dim \bm m_{k-1,i,c} < d$, use $G_c$ for $F_c$, from \eqref{eq:growingstatespace1}, and $R_c$ for $Q_c$, from \eqref{eq:growingstatespace2}, in the above.

\subsection{Associations}
\label{ss:associations}

The association sampling is the most involved part of the particle filter. In particular, the particle filter involves sampling the survival of each track, the number of new tracks, and the associations, at each time step. As has been seen in Subsection \ref{ss:trackdel}, sampling track survival is trivial, and the number of new tracks is inferred from the sampled associations. The reason the associations need to be sampled at all is that there are prohibitively many possible joint associations to consider each one; the number of options grows exponentially with each time step. It is with this in mind that we choose first to cluster the data deterministically. Doing so means that convergence of the particle filter to the true posterior is no longer guaranteed\footnote{In fact, as the prior is effectively changed, the particle filter converges to a different, but incredibly similar, posterior.}, but it is an acceptable approximation since it \textit{significantly} reduces the number of particles required for the particle filter to resemble accurately the distribution to which it converges. The alternatives are either to obtain the clusters probabilistically, which makes computation of the particle weight extremely difficult, or to not use clustering at all, which leads to useful information being ignored and many more particles being required. To do this, an estimate, $\bar{s'}_{\!\!k}^2$, of the observation variance is required. It is derived using all particles. Namely, we choose
\cm{
\bar{s'}_{\!\!k}^2 = \sum_{j=1}^J w_{k-1}^j \frac{b_{k-1}^j}{\varphi_{k-1}^j-1},
}{\label{eq:varestforallparts}}
where $w_{k-1}^j$ is the particle weight from the previous time step, and $b_{k-1}^j$ and $\varphi_{k-1}^j$ are the priors for $b$ and $\varphi$ at $k$ for particle $j$.

\subsubsection{Clustering}
\label{sss:clustering}

This clustering is done using a Kernel Smoothing Density (KSD) estimate over the data. Let
\cm{
f_\text{ksd}\left(z ~\big|~ \bm y_k,\bar{s'}_{\!\!k}^2\right) = \frac1{n_k}\sum_{l=1}^{n_k} \mathcal N\left(z ~\big|~ y_{kl},\bar{s'}_{\!\!k}^2\right).
}{\label{eq:ksd}}
Sufficiently close data should result in a so-called `combined peak'. As such, $f_\text{ksd}$ is numerically maximised $n_k$ times, starting from each datum $z=y_{kl}$, for $l\in\{1,\dots,n_k\}$. If the maxima resulting from the optimisation initialised by any $y_{kl}$ and $y_{kl'}$ are located approximately equally, then $y_{kl}$ and $y_{kl'}$ will be in the same cluster. 

Let $\mathcal G_k$ be the current set of $\Theta_k$ clusters at $k$, which gets initialised as $\mathcal G_k = \emptyset$, and let the argmax resulting from optimising from $y_{kl}$ be $z_{kl}$. Then, define the `argmax mean' of cluster $\mathcal G_{k\theta}$ with size $n_{k\theta}$ as
\cm{
\bar z_{k\theta} = (n_{k\theta})^{-1}\!\!\!\!\!\!\!\!\sum_{\{l':y_{kl}\in\mathcal G_{k\theta}\}} \!\!\!\!\!\!\!\! z_{kl'}.
}{\label{eq:argmaxmean}}
Working through the data sequentially, if $z_{kl}\approx \bar z_{k\theta}$ for some $\theta\le\Theta_k$, then update $\mathcal G_{k\theta}\leftarrow \mathcal G_{k\theta}\cup \{y_{kl}\}$. If $z_{kl}\not\approx \bar z_{k\theta}$ for any $\theta\le\Theta_k$, then begin a new cluster, $\mathcal G_{k,\Theta_k+1} = \{y_{kl}\}$, and update $\mathcal G_k \leftarrow \mathcal G_k \cup \{\mathcal G_{k,\Theta_k+1}\}$ and $\Theta_k \leftarrow \Theta_k + 1$.

For high $n_k$, this may become inefficient, in which case gating could be used to identify mutually separable groups of data (i.e., that could not appear in the same cluster) and optimise a separate KSD for each group starting from each datum in that group. In essence, this becomes a multi-fidelity clustering scheme, where larger clusters are crudely made to quickly rule out implausible options, and then more appropriate clusters are made in a more principled manner. Alternatively, the user can use any other clustering scheme they wish. 


\subsubsection{Sampling}
\label{sss:sampling}

Data in the same cluster will receive the same associations within each particle. For one particle (whose notation is suppressed for readability), define $\check A_{k\theta}$ as the \textit{initial} association in this particle of all data in $\mathcal G_{k\theta}$, for all $\theta\in\{1,\dots,\Theta_k\}$. This will be sampled from
\cm{
\check A_{k\theta} \sim q_{\check A}(\check A\mid \mathcal G_{k\theta},\bm\zeta_{1:k},\eta_{1:k-1},\bm a_{1:k-1},\bm y_{1:k-1}).
}{\label{eq:initassoc1}}
For $i\in\{1,\dots,N_{k-1}\}$, and with ${s'}_{\!\!k-1}^2 = \frac{b_{k-1}}{\varphi_{k-1}-1}$ and $\check\pi_{kic}=\pi_{k-1,i,c}$, we choose
\cm{
&q_{\check A}(\check A=i\mid\dots) \propto \zeta_{ki}\left(\frac{\alpha_{k-1,i}}{\beta_{k-1,i}}\right)^{n_{k\theta}}\\
&~\times\sum_{c=1}^{N_c}\!\mathcal N\left(\mathcal G_{k\theta}\mid\check m_{kic}\bm 1_{n_{k\theta}},\check V_{kic}\bm 1_{n_{k\theta}\times n_{k\theta}} \!+ {s'}_{\!\!k-1}^2I_{n_{k\theta}}\right)\check\pi_{kic},
}{\label{eq:initassoc2}}
for $i=0$, we select
\cm{
q_{\check A}(\check A=0\mid\dots) &\propto \left(\frac{\alpha_{k-1,0}}{\beta_{k-1,0}}\right)^{n_{k\theta}} \prod_{y\in\mathcal G_{k\theta}} p(y),
}{\label{eq:initassoc3}}
and, finally, for $i=N_{k-1}+1$, representing clusters coming from previously unseen tracks, we make
\cm{
q_{\check A}&(\check A=N_{k-1}+1\mid\dots)\\
&\propto 1(n_{k\theta}\ge2)\,\left(1 - \left(\frac{\xi_{k-1}}{1+\xi_{k-1}}\right)^{\varepsilon_{k-1}}\right)\,\left(\frac{\alpha_+}{\beta_+}\right)^{n_{k\theta}}\\
&~~~~\times\rho\left(\frac{2\pi{s'}_{\!\!k-1}^2}{n_{k\theta}}\right)^{\frac12}\prod_{y\in\mathcal G_{k\theta}}\mathcal N\left(y\mid \bar y_{k\theta},{s'}_{\!\!k-1}^2\right),
}{\label{eq:initassoc4}}
where $\bar y_{k\theta}:=\frac1{n_{k\theta}}\sum_{y\in\mathcal G_{k\theta}}y$ is a cluster mean. The intuition behind the various terms is as follows. For \eqref{eq:initassoc2} and \eqref{eq:initassoc3}, the $\frac{\alpha_{k-1,i}}{\beta_{k-1,i}}$ term is the expected number of data to be produced by track $i$, which is proportional to the prior probability that a given datum is associated to track $i$, hence its raising to the power of $n_{k\theta}$. The Gaussian and product terms represent the respective likelihoods of the proposed association, and, in \eqref{eq:initassoc2}, the $\zeta_{ki}$ ensures the probability of associating to inactive tracks is zero. In \eqref{eq:initassoc4}, the $\frac{\alpha_+}{\beta_+}$ plays the same role as its analogues did before, owing to the fact nothing has been learnt about these tracks yet (as they are unseen). Similarly, there is an (approximate) likelihood term at the end, from
\cm{
p&(\mathcal G_{k\theta}\mid \check A_{k\theta}=N_{k-1}+1)\\
&= \int p(\mathcal G_{k\theta}\mid X_{k,N_{k-1}+1})\,p(X_{k,N_{k-1}+1})\ dX_{k,N_{k-1}+1}\\
&\approx \rho \int \mathcal N\left(\mathcal G_{k\theta}\mid X\bm 1_{n_{k\theta}},{s'}_{\!\!k-1}^2I_{n_{k\theta}}\right)\ dX\\
&=\rho \left(\frac{2\pi{s'}_{\!\!k-1}^2}{n_{k\theta}}\right)^{\frac12}\prod_{y\in\mathcal G_{k\theta}}\mathcal N\left(y\mid \bar y_{k\theta},{s'}_{\!\!k-1}^2\right),
}{\label{eq:approxliknewtrack}}
where the approximation comes from using the full integral, rather than restricting it to the scene, $S$. This approximation is good whenever $\bar y_{k\theta}$ is not too close to the edge of the scene (relative to its estimated variance, ${s'}_{\!\!k-1}^2/n_{k\theta}$). This is particularly acceptable if the regions of interest are also not proximate with the edges of the scene. Returning to \eqref{eq:initassoc4}, the $1(n_{k\theta}\ge2)$ term forcibly prevents a `cluster' of size one initialising a new track. If preferred, this may be omitted. The most interesting term in \eqref{eq:initassoc4} is the $\left(1 - \left(\frac{\xi_{k-1}}{1+\xi_{k-1}}\right)^{\varepsilon_{k-1}}\right)$ term, which is the \textit{exact} prior probability that $\eta_k>0$.
Such a term is not required for the other possible $\check A_{k\theta}$, as cluster $\theta$ not being associated to a new track neither precludes nor ensures that $\eta_k>0$.

The final cluster associations, $A_{k\theta}$, are derived from the initial associations in the following way. Firstly, for all $\theta$ such that $\check A_{k\theta}\le N_{k-1}$, we set $A_{k\theta}=\check A_{k\theta}$. That is, data initially associated to pre-existing tracks (or clutter) go unchanged. This leaves distinguishing the new tracks from one another, which is done by sequentially sampling, allowing each cluster to either join previous clusters in a track, or start a new one. Consider sampling $A_{k\theta}$, by which point there are already $\check N_{k,\theta-1}$ new tracks. For $\iota \in\{1,\dots,\check N_{k,\theta-1}\}$, let
\cm{
\mathcal H'_{k\theta\iota} =\!\!\!\!\!\! \bigcup_{\{\theta'<\theta : A_{k\theta}=N_{k-1}+\iota\}}\!\!\!\!\!\!\!\!\!\!\!\!\!\!\!\!\mathcal G_{k\theta'}
}{\label{eq:newtrackcluster}}
be the $\nu'_{k\theta\iota}$ data already associated to the new track with index $N_{k-1}+\iota$, and let $\mathcal H_{k\theta\iota} = \mathcal H'_{k\theta\iota}\,\cup\,\mathcal G_{k\theta}$ also include the cluster under consideration, whose number of data is denoted $|\mathcal H_{k\theta\iota}|=\nu_{k\theta\iota} = \nu'_{k\theta\iota}+n_{k\theta}$. Also, recall $\bar y_{k\theta}=\frac1{n_{k\theta}}\sum_{y\in\mathcal G_{k\theta}}y$, and contrast it to the newly defined
\cm{
\bar y_{k\theta\iota} &:= \frac1{\nu_{k\theta\iota}}\sum_{y\in\mathcal H_{k\theta\iota}}\!\!\!y,&\bar y'_{k\theta\iota} &:= \frac1{\nu'_{k\theta\iota}}\sum_{y\in\mathcal H'_{k\theta\iota}}\!\!\!y.
}{\label{eq:allthesemeans}}
Now, we choose that, for all $\iota\in\{1,\dots,\check N_{k,\theta-1}+1\}$, the proposal mass function satisfies
\cm{
q_A&\left(A_{k\theta}=N_{k-1}+\iota ~\big|~ \{\mathcal G_{k\theta'},A_{k\theta'}\}_{\theta'=1}^{\theta-1},\mathcal G_{k\theta},\dots\right)\\
&\propto p\left(\mathcal G_{k\theta}~\big|~ A_{k\theta}=N_{k-1}+\iota, \{\mathcal G_{k\theta'},A_{k\theta'}\}_{\theta'=1}^{\theta-1},\dots\right)\\
&~~\times p\left(A_{k\theta}=N_{k-1}+\iota~\big|~ \{\mathcal G_{k\theta'},A_{k\theta'}\}_{\theta'=1}^{\theta-1},\dots\right)\\
&~~\times \left(\frac{\mathbb P(\eta_k=\check N_{k,\theta-1})}{\sqrt{n_{k\theta}}}\prod_{y\in\mathcal G_{k\theta}}\mathcal N\left(y\mid \bar y_{k\theta},{s'}_{\!\!k-1}^2\right)\right)^{-1}\\
}{\label{eq:finalassoc1}}
where the last line is an arbitrary \textit{constant} factor (with respect to $\iota$), facilitating a simpler and more interpretable proposal distribution. Then, for $\iota\le\check N_{k,\theta-1}$, \eqref{eq:finalassoc1} becomes
\cm{
q_A&\left(A_{k\theta}=N_{k-1}+\iota ~\big|~ \{\mathcal G_{k\theta'},A_{k\theta'}\}_{\theta'=1}^{\theta-1},\mathcal G_{k\theta},\dots\right)\\
&\propto \sqrt{\frac{n_{k\theta}\nu'_{k\theta\iota}}{n_{k\theta} + \nu'_{k\theta\iota}}}\exp\left\{-\frac{n_{k\theta} \nu'_{k\theta\iota} \left(\bar y_{k\theta} - \bar y'_{k\theta\iota}\right)^2}{2{s'}_{\!\!k-1}^2(n_{k\theta} + \nu'_{k\theta\iota})}\right\},
}{\label{eq:finalassoc1.5}}
using the results in Appendix \ref{app:gaussianratios}. Finally, we ask that, for $\iota=\check N_{k,\theta-1}+1$,
\cm{
q_A&\left(A_{k\theta}=N_{k-1}+\iota ~\big|~ \{\mathcal G_{k\theta'},A_{k\theta'}\}_{\theta'=1}^{\theta-1},\mathcal G_{k\theta},\dots\right)\\
&\propto \rho \sqrt{2\pi{s'}_{\!\!k-1}^2}\frac{\mathbb P(\eta_k=\iota)}{\mathbb P(\eta_k=\iota-1)}\\
&= \rho \sqrt{2\pi{s'}_{\!\!k-1}^2} \frac{\varepsilon_{k-1}+\check N_{k,\theta-1}}{\xi_{k-1}+1}.
}{\label{eq:finalassoc2}}
The (approximate) likelihoods in \eqref{eq:finalassoc1.5} (here, a ratio of the joint and marginal to obtain the conditional) and \eqref{eq:finalassoc2} come from \eqref{eq:approxliknewtrack}. 
It is interesting to observe that this proposal distribution takes care to consider the data jointly, yet the final probability for each cluster joining a track only depends on the cluster size and its mean, and the same quantities of the data currently comprising the new track. Finally, the association of each datum is simply inherited from the association of its cluster, so $a_{kl} = A_{k\theta_{kl}}$, where $\theta_{kl}$ is such that $y_{kl}\in\mathcal G_{k\theta_{kl}}$.

\subsection{Update}
\label{ss:update}

Both the tracks and hyperparameters require updating. As soon as the associations, $\bm a_k$, have been sampled our knowledge of $\gamma$, $s^2$ and all of the $\mu_i$, for $i\in\{0,\dots,N_{k-1}\}$ can be updated, as discussed in \ref{ss:hyperpars}, with the results in \eqref{eq:updategamma2}, \eqref{eq:updateinvgamma2} and \eqref{eq:updategamma3}. Then, for each of these tracks, the update is performed separately for each class, giving $\bm m_{kic}$ and $V_{kic}$ for each track, $i$, and class, $c$. The formulae for track $i\le N_{k-1}$, based on \eqref{eq:nhpp2}, given $\check{\bm m}_{kic}$ and $\check V_{kic}$ from \eqref{eq:trackpred1}, are
\cm{
\bm m_{kic} &= \check{\bm m}_{kic} + \frac{n_{ki}\left(\bar y_{ki}-(\check{\bm m}_{kic})_1\right)}{n_{ki}(\check V_{kic})_{1,1}+{s'}_{\!\!k}^2}(\check V_{kic})_{:,1},\\
V_{kic} &= \check V_{kic} - \frac{n_{ki}}{n_{ki}(\check V_{kic})_{1,1}+{s'}_{\!\!k}^2}(\check V_{kic})_{:,1}(\check V_{kic})_{1,:},
}{\label{eq:classupdate1}}
where $\bm y_{ki}=\{y_{kl}\mid a_{kl}=i\}$ are the data associated to track $i$, with mean $\bar y_{ki}$, ${s'}_{\!\!k}^2 = \frac{b_k}{\varphi_k-1}$ is the updated estimate of $s^2$, and, for example, $(\check V_{kic})_{:,1}$ is the first column of $\check V_{kic}$. These simplified results use \eqref{eq:typicalkalmanupdate1}-\eqref{eq:typicalkalmanupdate4}, in Appendix \ref{app:offsetdiag}. Alongside these, the class weights for each track, $\bm\pi_{ki}$, can be updated as described in \eqref{eq:updatedpi1} and \eqref{eq:updatedpi2}. 

For newly established tracks, for which the birth time is set as the current time $\check\kappa_i = k$, since the object's class effects only movement, no inference can yet be garnered, thus all class probabilities remain as the prior baseline, $\bm\pi_{ki} = \bm\pi_+$. For pre-existing tracks, each class has a separate update, but, here, they are all the same, so only one initial state mean and variance is required, coming from the initial distribution over the new track location. That distribution is
\cm{
p(X_{ki}\mid \bm y_{ki}) 
&\propto \prod_{y\in\bm y_{ki}} \mathcal N\left(y~\big|~ X_{ki},{s'}_{\!\!k}^2\right)\,1(X_{ki}\in S)\\
&\approx \mathcal N\left(X_{ki}~\Big|~ \bar y_{ki},\frac{{s'}_{\!\!k}^2}{n_{ki}}\right).
}{\label{eq:initstate1}}
Recall that, initially, the state vector is simply $\bm x_{ki} = (X_{ki})\in\mathbb R^1$. As such, the initial state distribution is 
\cm{
\bm x_{ki}\mid \bm y_{ki} \sim &~\mathcal N(\bm m_{ki},V_{ki}),\\
\bm m_{ki} = \bar y_{ki}\bm 1_1,~~~&V_{ki} = \frac{{s'}_{\!\!k}^2}{n_{ki}}\bm 1_{1,1}.
}{\label{eq:initstate2}}
For completeness, define $\bm m_{kic}=\bm m_{ki}$ and $V_{kic}=V_{ki}$, for all $c$ and new track indices, $i$. As previously mentioned, the status of new tracks is guaranteed to be active, so $\zeta_{ki}=1$ is known. Finally, the prior count parameters are taken as the baseline values, as mentioned in Subsection \ref{ss:hyperpars}, but they can be immediately updated as
\cm{
\alpha_{ki} &= \alpha_+ + n_{ki},&\beta_{ki} &= \beta_+ + 1.
}{\label{eq:updatemubaseline}}

\subsection{Particle Weights}
\label{ss:weights}

The above is performed for each of the $j\le J$ particles, yielding $\{\eta_{1:k}^j,\bm\zeta_{1:k}^j,\bm a_{1:k}^j\}_{j=1}^J$. Each of these has an associated weight, $w_k^j$, to be computed such that the weighted samples approximate the true joint posterior distribution. From e.g., \cite{doucet2009}, it is known that 
\cm{
w_k^j &\propto n_k!u_k^j\,w_{k-1}^j,\\
u_k^j &= p(\bm y_k\mid \eta_{1:k}^j,\bm\zeta_{1:k}^j,\bm a_{1:k}^j,\bm y_{1:k-1})\\
&~~\times \frac{p(\eta_k^j,\bm\zeta_k^j,\bm a_k^j\mid \eta_{1:k-1}^j,\bm\zeta_{1:k-1}^j,\bm a_{1:k-1}^j,\bm y_{1:k-1})}{q(\eta_k^j,\bm\zeta_k^j,\bm a_k^j\mid \eta_{1:k-1}^j,\bm\zeta_{1:k-1}^j,\bm a_{1:k-1}^j,\bm y_{1:k})},
}{\label{eq:weightincdefinition}}
where the $n_k!$ comes from the proportionality, for later convenience. Now define
\cm{
P_k^j &:= p(\bm y_k\mid \eta_{1:k}^j,\bm\zeta_{1:k}^j,\bm a_{1:k}^j,\bm y_{1:k-1}),\\
\check P_k^j &:= p(\eta_k^j,\bm\zeta_k^j,\bm a_k^j\mid \eta_{1:k-1}^j,\bm\zeta_{1:k-1}^j,\bm a_{1:k-1}^j,\bm y_{1:k-1}),\\
Q'{}_{\!\!k}^j &:= q(\eta_k^j,\bm\zeta_k^j,\bm a_k^j\mid \eta_{1:k-1}^j,\bm\zeta_{1:k-1}^j,\bm a_{1:k-1}^j,\bm y_{1:k}).
}{\label{eq:weightinccomponents}}
Because the sampling scheme is unique -- there is only one way to obtain any given sample, $\{\eta_k,\bm\zeta_k,\bm a_k\}$ -- $Q'{}_{\!\!k}^j$ can be recorded during the sampling process. Specifically, if
\cm{
\!\!\!\!\check Q'{}_{\!\!k}^j &= q(\eta_k^j, \bm a_k^j \mid \eta_{1:k-1}^j, \bm\zeta_{1:k}^j, \bm a_{1:k-1}^j, \bm y_{1:k})\\
&= \prod_{\theta=1}^{\Theta_k} q_A(A_{k\theta}^j \mid \{\mathcal G_{k\theta'}, A_{k\theta'}^j\}_{\theta'=1}^{\theta-1}, \mathcal G_{k\theta},\dots)\\
&~~~~~~~~\times q_{\check A}(\check A_{k\theta}^j\mid \mathcal G_{k\theta}, \bm\zeta_{1:k}^j, \eta_{1:k-1}^j, \bm a_{1:k-1}^j, \bm y_{1:k-1}),
}{\label{eq:qdecomp1}}
is computed when sampling $\{\check A_{k\theta}^j,A_{k\theta}^j\}_{\theta=1}^{\Theta_k}$, then the final proposal probability is
\cm{
Q'{}_{\!\!k}^j &= \check Q'{}_{\!\!k}^j \cdot q(\bm\zeta_k^j \mid \eta_{1:k-1}^j, \bm\zeta_{1:k-1}^j, \bm a_{1:k-1}^j, \bm y_{1:k})\\
&= \check Q'{}_{\!\!k}^j \cdot \prod_{i=1}^{N_{k-1}^j} (1-\psi)^{\check\zeta_{ki}^j(1-\zeta_{ki}^j)}\,\psi^{\zeta_{ki}^j}.
}{\label{eq:qdecomp2}} 
What remains is to compute the likelihood and prior terms, $P_k^j$ and $\check P_k^j$, respectively. Starting with the former, we have
\cm{
P_k^j &= p(\bm y_{k0}^j)\prod_{i=1}^{N_k^j} p(\bm y_{ki}^j\mid \eta_{1:k}^j,\bm\zeta_{1:k}^j,\bm a_{1:k}^j,\bm y_{1:k-1},\{\bm y_{k\iota}^j\}_{\iota=1}^{i-1}).
}{\label{eq:weightlikelihood0}}
Each factor in the product regarding pre-existing objects can be expressed as a weighted sum of marginal likelihoods, weighted by $\bm\pi_k^j$. Namely, for $i\in\{1,\dots,N_{k-1}^j\mid \zeta_{ki}^j=1\}$, we have
\cm{
p&(\bm y_{ki}^j\mid \eta_{1:k}^j,\bm\zeta_{1:k}^j,\bm a_{1:k}^j,\bm y_{1:k-1},\{\bm y_{k\iota}^j\}_{\iota=1}^{i-1})\\
&= \sum_{c=1}^{N_c} \mathcal N\Big(\bm y_{ki}^j~\big|~(\check{\bm m}_{kic}^j)_1\bm 1_{n_{ki}^j},\\
&~~~~~~~~~~~~~~~~~~~(\check V_{kic}^j)_{1,1}\bm 1_{n_{ki}^j\times n_{ki}^j}+{s'}_{\!\!k}^{j2} I_{n_{ki}^j}\Big)\,\check\pi_{kic}^j.
}{\label{eq:weightlikelihood0.1}}
For the corresponding factors for $i\in\{N_{k-1}^j+1,\dots,N_k^j\}$ one can use the approximation in \eqref{eq:approxliknewtrack} to obtain
\cm{
p(\bm y_{ki}^j&\mid \eta_{1:k}^j,\bm\zeta_{1:k}^j,\bm a_{1:k}^j,\bm y_{1:k-1},\{\bm y_{k\iota}^j\}_{\iota=1}^{i-1})\\
&\approx \rho\sqrt{\frac{2\pi{s'}_{\!\!k}^{j2}}{n_{ki}^j}} \mathcal N\left(\bm y_{ki}^j~\big|~\bar y_{ki}^j\bm 1_{n_{ki}^j},{s'}_{\!\!k}^{j2}I_{n_{ki}^j}\right).
}{\label{eq:weightlikelihood0.2}}
The joint likelihoods can be efficiently computed using \eqref{eq:simplifylikelihood1} and \eqref{eq:simplifylikelihood2}, from Appendix \ref{subapp:uses}. 

Finally, consider
\cm{
n_k!\,\check P_k^j &= n_k!\,p(\bm a_k^j\mid \eta_{1:k}^j,\bm\zeta_{1:k}^j,\bm a_{1:k-1}^j,\bm y_{1:k-1})\\
&~~~\times p(\bm\zeta_k^j\mid \eta_{1:k}^j,\bm\zeta_{1:k-1}^j,\bm a_{1:k-1}^j,\bm y_{1:k-1})\\
&~~~\times p(\eta_k^j\mid \eta_{1:k-1}^j,\bm\zeta_{1:k-1}^j,\bm a_{1:k-1}^j,\bm y_{1:k-1}).
}{\label{eq:weightprior1}}
Combinatorially, the association mass function evaluated at $\bm a_k^j$ can be computed as the product of the probabilities of having $n_k$ data in total, and these specific (ordered) track associations, marginalised over the Poisson means, $\bm\mu_k^j := (\mu_0,\dots,\mu_{N_k^j})^T$. That is,
\cm{
\!n_k!&p(\bm a_k^j\mid \eta_{1:k}^j,\bm\zeta_{1:k}^j,\bm a_{1:k-1}^j,\bm y_{1:k-1})\\
&= n_k!\!\int \text{Poisson}(n_k\mid\check\mu_k^j)\\
&~~~~~~~~\times\prod_{i=0}^{N_k^j} \left(\frac{\mu_i}{\check\mu_k^j}\right)^{n_{ki}^j} \!\!\!\! \text{Gamma}(\mu_i \mid \alpha_{k-1,i}^j, \beta_{k-1,i}^j) \ d\bm\mu_k^j\\
&= \!\!\!\!\prod_{\substack{i=0\\i:\zeta_{ki}=1}}^{N_k^j} \frac{\Gamma(\alpha_{k-1,i}^j+n_{ki}^j)}{\Gamma(\alpha_{k-1,i}^j)}\frac{{\beta_{k-1,i}^j}^{\alpha_{k-1,i}^j}}{(\beta_{k-1,i}^j+1)^{\alpha_{k-1,i}^j+n_{ki}^j}},
}{\label{eq:condassocprob}}
with the sum of the means of active tracks $\check\mu_k^j:=\mu_0 + \sum_{i=1}^{N_k^j}\mu_i\zeta_{ki}^j$, and where we say that $\zeta_{k0}=1$, for notational brevity. It is easily shown that
\cm{
p(\bm\zeta_k^j&\mid \eta_{1:k}^j,\bm\zeta_{1:k-1}^j,\bm a_{1:k-1}^j,\bm y_{1:k-1})\\
&= \prod_{i=1}^{N_{k-1}^j} (1-\psi)^{\zeta_{k-1,i}^j(1-\zeta_{ki}^j)} \, \psi^{\zeta_{ki}^j},
}{\label{eq:weightprior1.1}}
as well as
\cm{
p(\eta_k^j&\mid \eta_{1:k-1}^j,\bm\zeta_{1:k-1}^j,\bm a_{1:k-1}^j,\bm y_{1:k-1})\\
&= \frac{\Gamma(\varepsilon_{k-1}^j+\eta_k^j)}{\Gamma(\varepsilon_{k-1}^j)\eta_k^j!}\frac{{\xi_{k-1}^j}^{\varepsilon_{k-1}^j}}{(\xi_{k-1}^j+1)^{\varepsilon_{k-1}^j+\eta_k^j}},
}{\label{eq:weightprior1.2}}
which completes the computation of $n!\check P_k^j$. For convenience, the product term in \eqref{eq:qdecomp2} and the result in \eqref{eq:weightprior1.1} can be combined to
\cm{
\frac1{Q'{}_{\!\!k}^j}&p(\bm\zeta_k^j\mid \eta_{1:k}^j,\bm\zeta_{1:k-1}^j,\bm a_{1:k-1}^j,\bm y_{1:k-1})\\
&~~~~~~~~= \frac1{\check Q'{}_{\!\!k}^j}\prod_{i=1}^{N_{k-1}^j} \frac{(1-\psi)^{\zeta_{k-1,i}^j(1-\zeta_{ki}^j)} \, \psi^{\zeta_{ki}^j}}{(1-\psi)^{\check\zeta_{ki}^j(1-\zeta_{ki}^j)} \, \psi^{\zeta_{ki}^j}}\\
&~~~~~~~~= \frac1{\check Q'{}_{\!\!k}^j}\prod_{i=1}^{N_{k-1}^j} (1-\psi)^{(\zeta_{k-1,i}^j-\check\zeta_{ki}^j)(1-\zeta_{ki}^j)}.
}{\label{eq:weightpriorsimplification}}

This completes GaPP-Class, given in Algorithm \ref{alg:gappclass}. An optional particle resampling step (see e.g.,\cite{sarkka2013}) is added to reduce degeneracy.
\begin{algorithm} 
    \caption{GaPP-Class at step $k$}
    \begin{algorithmic}
        \Require data $\bm y_k$, all particle quantities and weights up to $k-1$
        \State resample particles
        \State cluster data (sec. \ref{sss:clustering})
        \For{particle $j=1,\dots,J$}
            \For{pre-existing tracks}
                \State perform initial and final track deletion (sec. \ref{ss:trackdel})
                \State do predict step, if track still active (sec. \ref{ss:trackprediction})
            \EndFor
            \State sample data association $\bm a_k^j$ (sec. \ref{sss:sampling})
            \State update NHPP and birth rate hyperparameters (sec. \ref{ss:update})
            \For{pre-existing tracks}
                \State do update step for pre-existing tracks with \eqref{eq:classupdate1}
                \State update class probabilities with \eqref{eq:updatedpi1}--\eqref{eq:updatedpi2}
            \EndFor
            \State initialise new tracks with \eqref{eq:initstate2}--\eqref{eq:updatemubaseline}
            \State calculate weight increment (sec. \ref{ss:weights})
        \EndFor
        \State normalise weights $w_k^j$
        \Ensure all particle quantities and weights up to $k$
    \end{algorithmic}
    \label{alg:gappclass}
\end{algorithm}

\section{Revival MCMC Kernel}
\label{sec:revival}

GaPP-Class is \textit{extended} into GaPP-ReaCtion (with minimal modification), and so the output (at time step $k$) of GaPP-Class is the input for this revival-based extension, where new tracks can be reclassified as pre-existing tracks which were prematurely terminated. This takes the form of a MCMC kernel, specifically of the Metropolis-within-Gibbs \cite{hastings1970, geman1984, gelfand2000gibbs} variety. 
The outputs of GaPP-Class will be the \textit{first} current sample of GaPP-ReaCtion, so these outputs 
are now denoted with a `hat' (e.g., $\hat{\bm m}_{kic}$) if preliminary and `breve' (e.g., $\breve{\bm m}_{kic}$) if updated, instead of `check' (e.g., $\check{\bm m}_{kic}$) and no accent (e.g., $\bm m_{kic}$), respectively. This kernel operates on the sample by sequentially considering every track capable of having its activity status changed: revival is having $\zeta_{ki}$ change from 0 to 1, and splitting is the reverse. For each of these steps, there is a current sample (using `hat' and `breve' accents), the proposed sample (using `tilde', e.g., $\tilde\eta_k$, for updated quantities), the possible samples which could have been proposed (using `tildes' and superscripts where relevant), and the final sample (again using `check' accents from preliminary variables and no accents for updated variables). The final sample is the proposed sample, if accepted, and the current sample otherwise. 
Since this is a Metropolis-Hastings MCMC kernel, any accepted sample must have a non-zero probability of proposing the current sample. This takes the form of allowing an active track to be split, retroactively deleted at some recent time, and for a new track to begin at the current time.

We now specify the conditions of a revival. Clearly, as the uncertainty of an object grows the longer it survives undetected, there must be some expiry time on a track's ability to be revived, which we denote $d_\zeta$ (common for all tracks). Else, any new track could arbitrarily be assigned to a previously terminated one without meaningful evidence. Also, the additional heuristic track termination methods are absolute; not only would one not want to revive a track that was, say, explicitly deleted because the variance was already too high, but allowing its revival would result in a sample that could not be attained by the GaPP-Class particle filter section. The reason these deletion criteria are included at all is to exclude them from the support of the particle filter. Allowing these criteria to be circumvented by revival would lead to the revival kernel targeting a different posterior to that of the particle filter. 

Let $\Omega_{ki}\in\mathbb Z_2$ be the validity of a track to be revived. (This is in the \textit{final} sample; $\breve\Omega_{ki}$ and $\tilde\Omega_{ki}$ are the analogues in the current and proposed samples, respectively.) Also, let 
\cm{
\kappa'_{ki} = \max \{\kappa<k\mid |\{l\mid a_{\kappa l}=i\}|>0\} \cup \{k-d_\zeta\}.
}{\label{eq:lastinvalidrevivaltime}}
Then, if, at step $k$, object $i$ has been deleted within the last $d_\zeta - (k-\kappa'_{ki})$ time steps (including at $k$) via sampling, then $\Omega_{ki} := 1$, so revival is valid. Otherwise, $\Omega_{ki} := 0$ and one of the following statements is true: `object $i$ is still active', `object $i$ was deleted via the heuristic criteria', or `object $i$ was deleted by sampling at step $\kappa_i \le 2k-d_\zeta - \kappa'_{ki}$, i.e., more than $d_\zeta - (k-\kappa'_{ki})$ steps ago'. We now define the proposal process for taking a newly initialised track and attaching it to a previously deleted track, and then the proposal process for splitting an active track. 

\subsection{Revival Proposal}
\label{ss:revivalproposal}

Consider the revival proposal for a newly initialised track with index $\iota$ being the continuation of the terminated track with index $i$. For the probability of this proposal to be non-zero, the \textit{current} sample must have $2k-d_\zeta - \breve\kappa'_{ki} < \breve \kappa_i \le k = \hat\kappa_\iota$ and have had track $i$ be deleted via sampling, thus making $\breve\Omega_{ki}=1$, and hence making this a valid revival. Defining exactly what a revival is is equivalent to defining the differences between the samples directly. With the current sample being $\{\breve{\bm a}_{1:k}, \breve{\bm\zeta}_{1:k}, \breve\eta_{1:k}\}$, then, were we to propose attaching track $\iota$ to track $i$, the proposed sample would be $\{\tilde{\bm a}_{1:k}, \tilde{\bm\zeta}_{1:k}, \tilde\eta_{1:k}\}$, satisfying
\cm{
\tilde{\bm a}_{1:k-1} &= \breve{\bm a}_{1:k-1}, & \tilde{\bm\zeta}_{1:\breve\kappa_i-1} &= \breve{\bm\zeta}_{1:\breve\kappa_i-1}, & \tilde\eta_{1:k-1} &= \breve\eta_{1:k-1},
}{\label{eq:proposedsamp1}}
and
\cm{
\tilde a_{kl} &= 
\begin{cases}
    \breve a_{kl},& \breve a_{kl}\ne\iota\\
    i,& \breve a_{kl} = \iota
\end{cases},~ & \tilde\eta_k &= \breve\eta_k - 1\\
\tilde\zeta_{k'i'} &= 
\begin{cases}
    \breve\zeta_{k'i'},& i'\ne i\\
    1,& i' = i
\end{cases},~ & \forall\,k'&\in\{\breve\kappa_i,\dots,k\}.
}{\label{eq:proposedsamp2}}
It should be noted that $\tilde\zeta_{k'\iota}$ is undefined, since there is no such object in the proposed scenario. The other possible proposal is to simply leave $\iota$ as a new track, in which case $\{\tilde{\bm a}_{1:k}, \tilde{\bm\zeta}_{1:k}, \tilde\eta_{1:k}\} = \{\breve{\bm a}_{1:k}, \breve{\bm\zeta}_{1:k}, \breve\eta_{1:k}\}$ because there is no change. We write the proposal mass function as $q_r(i\mid\iota)$ as a shorthand for $q(\tilde{\bm a}_{1:k}^{i\iota}, \tilde{\bm\zeta}_{1:k}^{i\iota}, \tilde\eta_{1:k}^{i\iota}\mid \breve{\bm a}_{1:k}, \breve{\bm\zeta}_{1:k}, \breve\eta_{1:k}, \bm y_{1:k})$, using the superscripts to distinguish between the possible proposals and the one sampled. The option of not joining track $\iota$ is denoted $q_r(0\mid\iota)$.

We choose the proposal distribution as
\cm{
q_r&(i\mid\iota) \propto \frac{p(\tilde{\bm a}_{1:k}^{i\iota}, \tilde{\bm\zeta}_{1:k}^{i\iota}, \tilde\eta_{1:k}^{i\iota}\mid \bm y_{1:k})}{p(\breve{\bm a}_{1:k}, \breve{\bm\zeta}_{1:k}, \breve\eta_{1:k} \mid \bm y_{1:k})}.
}{\label{eq:qr1}}
Axiomatically, we have $q_r(0\mid\iota)\propto 1$. Then, for valid $i\ne0$, the differences between the proposed and current scenarios are that track $i$ survived between time steps $\breve\kappa_i$ and $k$ (inclusive), rather than being deleted at $\breve\kappa_i$. As such, it actively associated zero data between $\breve\kappa_i$ and $k-1$ (inclusive). There is one fewer new track at $k$ and only the likelihood of the data $\breve{\bm y}_{k\iota}=\{y_{kl}\mid \breve a_{kl} = \iota\}$ is now different. Given the other quantities which are common to both samples, there are no other terms required in \eqref{eq:qr1}. Thus, after sufficient algebraic simplification, we have
\cm{
q_r&(i\mid\iota)\propto \frac{p(\breve{\bm y}_{k\iota}\mid \tilde{\bm a}_{1:k}^{i\iota}, \tilde{\bm\zeta}_{1:k}^{i\iota}, \tilde\eta_{1:k}^{i\iota},\bm y_{1:k} \setminus \breve{\bm y}_{k\iota})}{p(\breve{\bm y}_{k\iota}\mid \breve{\bm a}_{1:k}, \breve{\bm\zeta}_{1:k}, \breve\eta_{1:k}, \bm y_{1:k} \setminus \breve{\bm y}_{k\iota})}\\
&~\times \frac{\Gamma(\breve\alpha_{\breve\kappa_i-1,i}+\breve n_{k\iota})\breve\beta_{\breve\kappa_i-1,i}^{\breve\alpha_{\breve\kappa_i-1,i}}}{\Gamma(\breve\alpha_{\breve\kappa_i-1,i})(\breve\beta_{\breve\kappa_i-1,i}+k-\breve\kappa_i+1)^{\breve\alpha_{\breve\kappa_i-1,i}+\breve n_{k\iota}}}\\
&~\times \frac{\Gamma(\alpha_+)(\beta_++1)^{\alpha_++\breve n_{k\iota}}}{\Gamma(\alpha_++\breve n_{k\iota})\beta_+^{\alpha_+}}\frac{\breve\eta_k(\breve\xi_{k-1}+1)}{\breve\varepsilon_{k-1}+\breve\eta_k-1}\frac{\psi^{k-\breve\kappa_i+1}}{1-\psi},
}{\label{eq:qr2}}
where $\bm y_{1:k}\setminus \breve{\bm y}_{k\iota}$ are all data except $\breve{\bm y}_{k\iota}$. The likelihood terms are computed as in \eqref{eq:weightlikelihood0.1} and \eqref{eq:weightlikelihood0.2}.

\subsection{Splitting Proposal}
\label{ss:splittingproposal}

A `split' is defined as retrospectively terminating an active track at a previous time step, and assigning the associated data at the current time step to a new track. The splitting distribution is very simplistic; for a given active track $i$ the splitting time is sampled uniformly over the valid time steps. The only potential issue requiring resolution is what the valid times are. Since this is the inverse to the revival proposal distribution, there must be associated data at the current time step, else there would be no new track with which to revive track $i$. Thus, we only consider a split of track $i$ if $|\{l\mid \breve a_{kl}=i\}|>0$. Then, because the revival track does not resample any associations, there must also be no associated data between the split time and step $k-1$ (inclusive). Revivals may only occur to tracks deleted within the last $d_\zeta$ time steps, and so the split time may not be more than $d_\zeta$ steps prior. We also choose that tracks which were revived at $k$ cannot be immediately re-split, but this can be omitted if desired. As such, recalling the definition of $\breve\kappa'_{ki}$ (described there for the \textit{final} sample, $\kappa'_{ki}$) from atop Subsection \ref{ss:revivalproposal} and with $q_s(\kappa\mid i)$ as the splitting proposal distribution, we can say that
\cm{
q_s(\kappa\mid i) &= \left(\frac1{k-\breve\kappa'_{ki}}\right)\ 1(\breve\kappa'_{ki}<\kappa\le k)\ 1(\breve\zeta_i=1),
}{\label{eq:qs}}
Contrary to the revival process, there is no probability of sampling `no split', in which the proposed sample is the same as the current. The reason the revival is not formulated in the same way is because this way provides a principled yet straightforward method for the current sample to essentially inform the probability of using the revival kernel at all, whereas the splitting kernel prioritises simplicity, so this probability is always one.

\subsection{Acceptance Probabilities}
\label{ss:acceptanceprobabilities}

\subsubsection{Revival Acceptance}
\label{sss:revivalacc}

Assume that, for a given new track $\iota$, sampled is $\tilde\imath\sim q_r(i\mid\iota)$, resulting in $\{\tilde{\bm a}_{1:k}, \tilde{\bm\zeta}_{1:k}, \tilde\eta_{1:k}\}$ being the proposed sample. If $\tilde\imath=0$, then no acceptance probability is needed, since the current and proposed samples are the same. So, consider the case in which $\tilde\imath\ne0$. The acceptance probability of this proposal is derived from the Metropolis acceptance probability,
\cm{
\Lambda_{k\iota\tilde\imath} = \min\{1,\lambda_{k\iota\tilde\imath}\},
}{\label{eq:accprobtechnicality}}
where
\cm{
\lambda_{k\iota\tilde\imath} 
&= \frac{p(\tilde{\bm a}_{1:k},\tilde{\bm\zeta}_{1:k},\tilde\eta_{1:k}\mid \bm y_{1:k})}{p(\breve{\bm a}_{1:k},\breve{\bm\zeta}_{1:k},\breve\eta_{1:k}\mid \bm y_{1:k})}\ \frac{q_s(\breve\kappa_{\tilde\imath}\mid\tilde\imath)}{q_r(\tilde\imath\mid\iota)},
}{\label{eq:revivalaccprob1}}
where the split time in $q_s$ is the time of deletion of $\tilde\imath$, which would require sampling were this revival to be reversed. In order to sample $\tilde\imath$, $q_r(i\mid\iota)$ was already computed for all $i$, in Subsection \ref{ss:revivalproposal}, including computing the normalising constant,
\cm{
\tilde Z_{k\iota} 
&= 1 + \!\!\!\!\sum_{\{i\mid \breve\Omega_{ki}=1\}} \frac{p(\tilde{\bm a}_{1:k}^{i\iota}, \tilde{\bm\zeta}_{1:k}^{i\iota}, \tilde\eta_{1:k}^{i\iota}\mid \bm y_{1:k})}{p(\breve{\bm a}_{1:k}, \breve{\bm\zeta}_{1:k}, \breve\eta_{1:k}\mid \bm y_{1:k})}.
}{\label{eq:revivalnormconstant}}
Now, the first fraction in \eqref{eq:revivalaccprob1} cancels with $q_r(\tilde\imath\mid\iota)$, leaving
\cm{
\lambda_{k\iota\tilde\imath} &= \frac{\tilde Z_{k\iota}}{k-\tilde\kappa'_{k\tilde\imath}},
}{\label{eq:revivalaccprob2}}
where $q_s(\breve\kappa_{\tilde\imath}\mid \tilde\imath) = (k - \tilde\kappa'_{k\tilde\imath})^{-1}$ is the inverse of the number of valid split times were $\tilde\imath$ being split, as the indicator terms in \eqref{eq:qs} are necessarily satisfied.

\subsubsection{Split Acceptance}
\label{sss:splitacc}

\begin{figure*}[tp]
    \centering
    \includegraphics[width=0.32\linewidth]{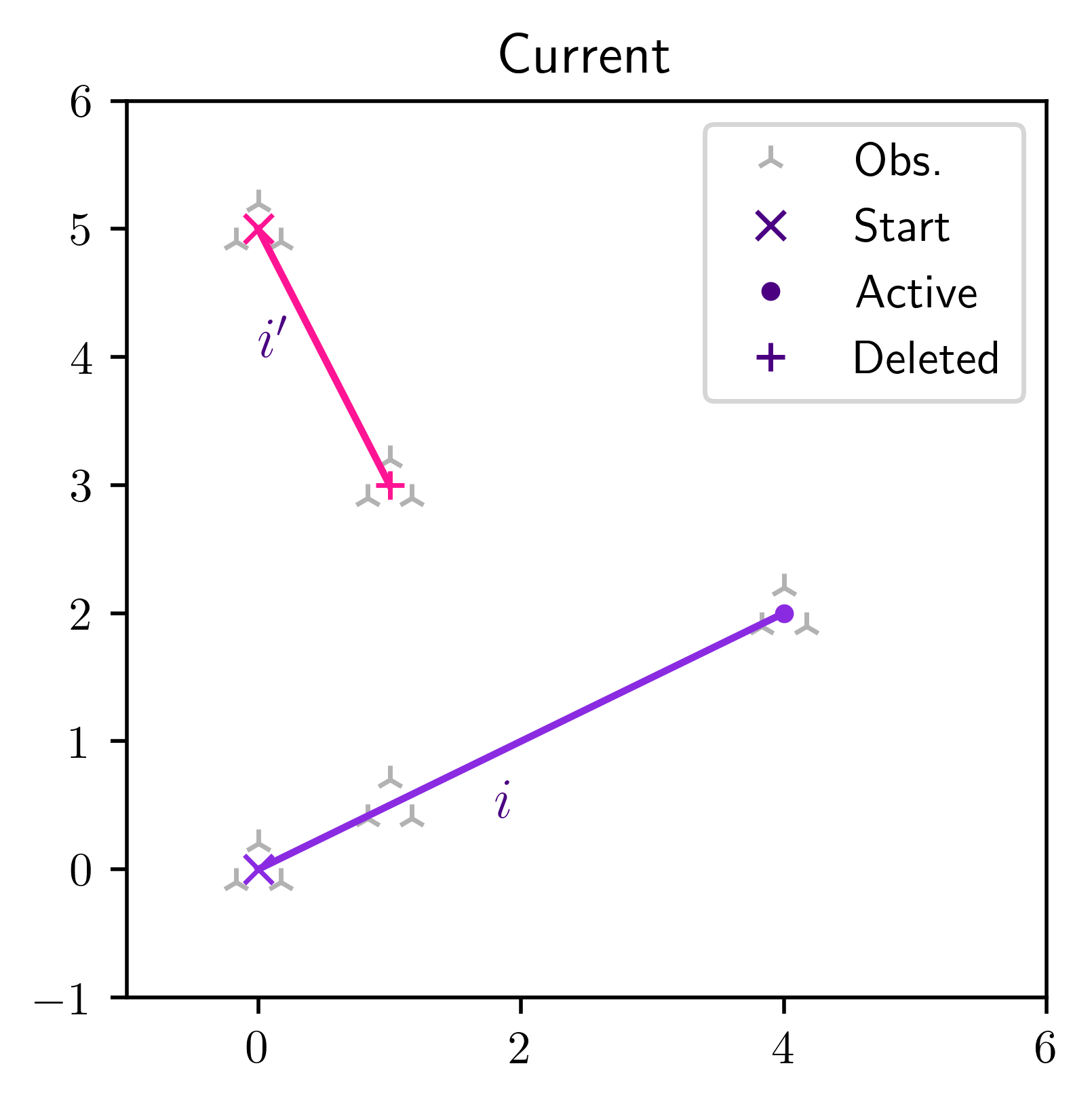}
    \includegraphics[width=0.32\linewidth]{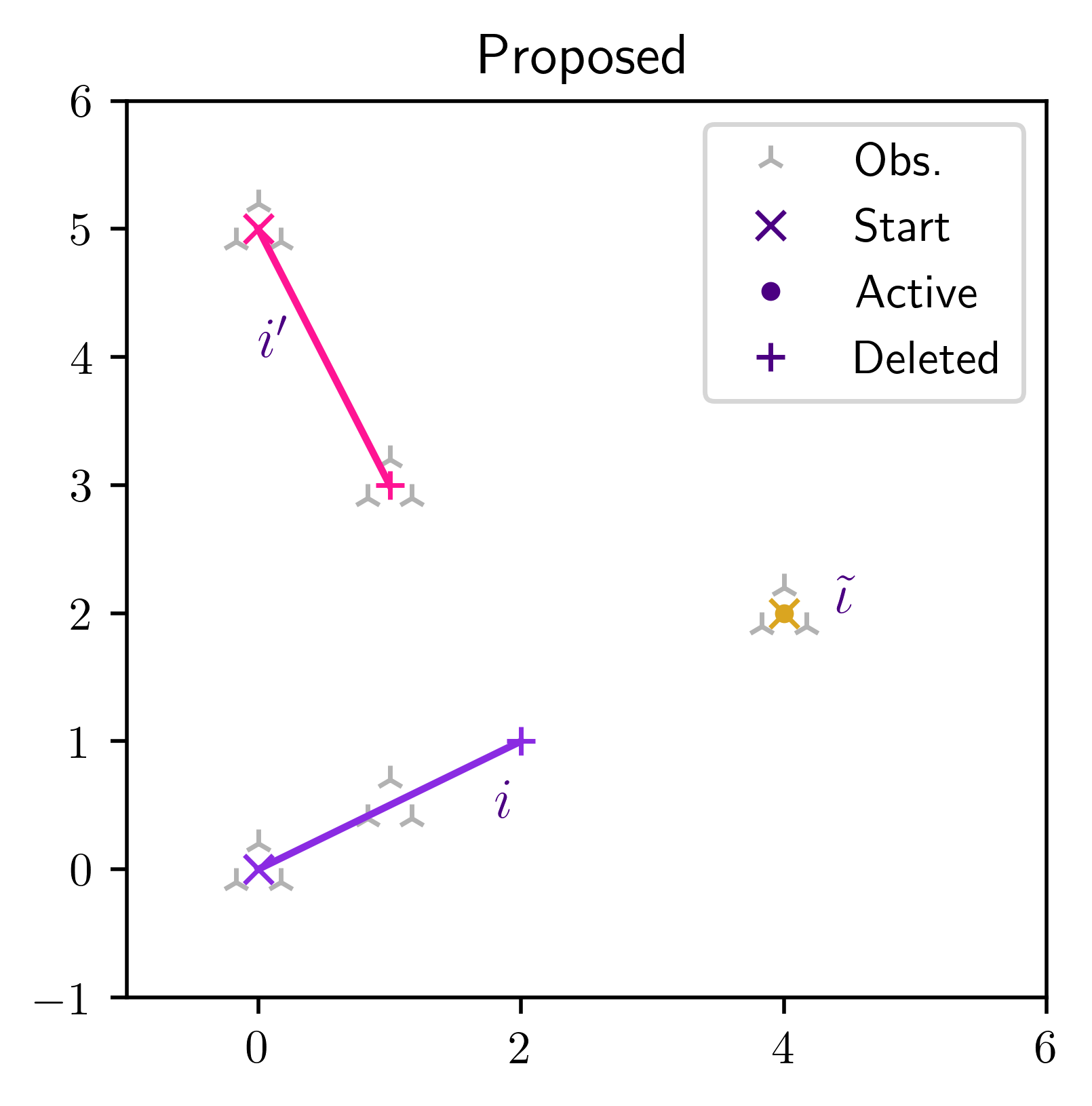}
    \includegraphics[width=0.32\linewidth]{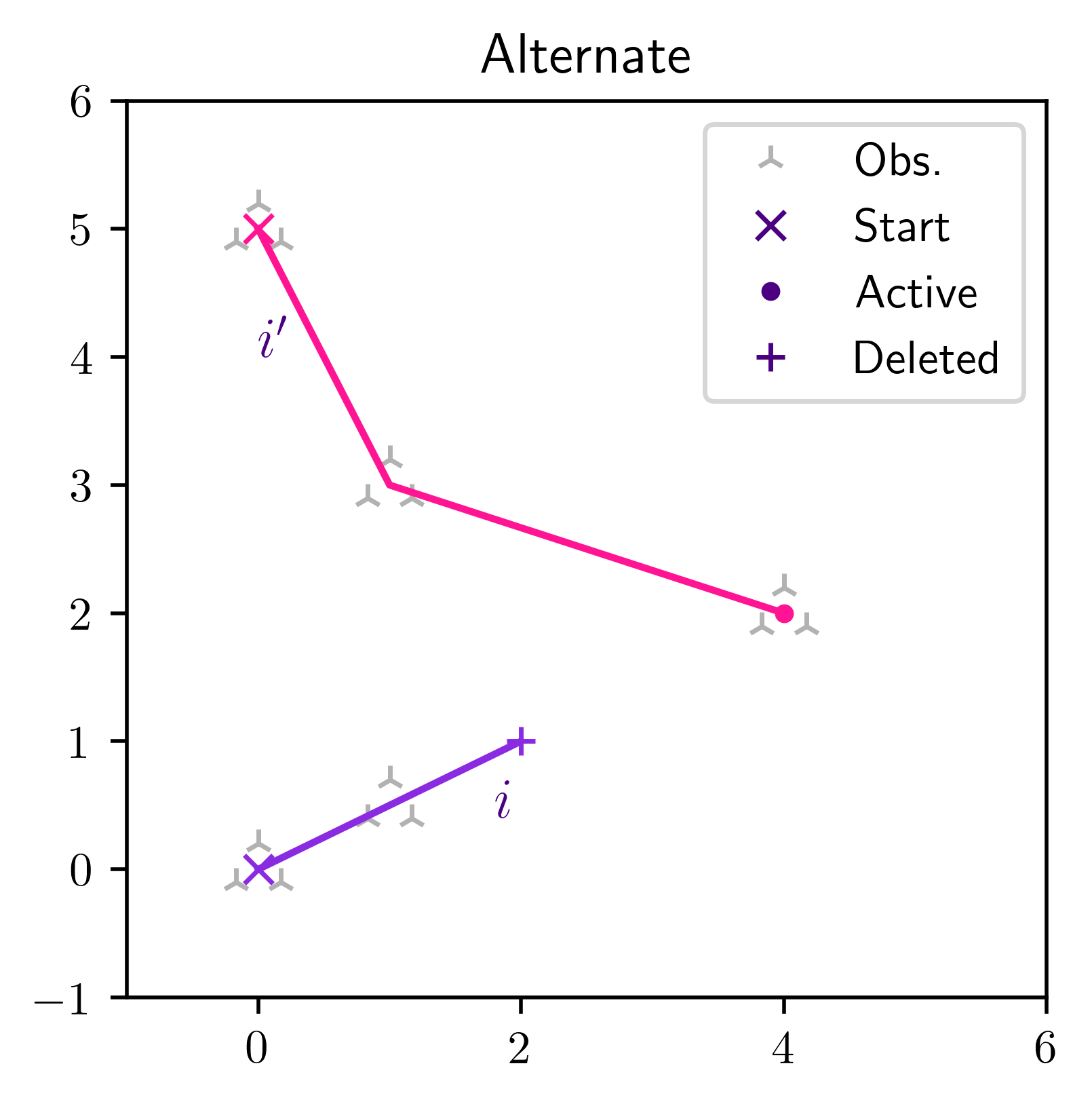}
    \caption{A toy example illustrating the types of sample relevant to the split acceptance probability. Left to right, the samples are the current ($\{\breve{\bm a}_{1:k}, \breve{\bm\zeta}_{1:k}, \breve\eta_{1:k}\}$), the proposed ($\{\tilde{\bm a}_{1:k}, \tilde{\bm\zeta}_{1:k}, \tilde\eta_{1:k}\}$), in which track $i$ is split creating $\tilde\iota$, and a non-inverse revival of the split ($\{\breve{\bm a}_{1:k}^{i'\tilde\iota},\breve{\bm\zeta}_{1:k}^{i'\tilde\iota},\breve\eta_{1:k}^{i'\tilde\iota}\}$), in which track $i'$ is revived using $\tilde\iota$. 
    }
    \label{fig:splitsampletypes}
\end{figure*}

Now assume, for a given active track $i$ with at least some associated data at $k$ (that is, $|\{l\mid \breve a_{kl}=i\}|>0$), that $\{\tilde{\bm a}_{1:k}, \tilde{\bm\zeta}_{1:k}, \tilde\eta_{1:k}\}$ represents the sampling of $\tilde\kappa\sim q_s(\kappa\mid i)$, which yields a new object whose label we denote $\tilde\iota$. Let the acceptance probability be $\Lambda'_{k\tilde\kappa i}=\min\{1, \lambda'_{k\tilde\kappa i}\}$. Then, by the definition in the first line of \eqref{eq:revivalaccprob1}, one can immediately see that 
\cm{
\lambda'_{k\tilde\kappa i} = \lambda_{k\tilde\iota i}^{-1} = \frac{k-\breve\kappa'_{ki}}{\breve Z_{k\tilde\iota}},
}{\label{eq:splitaccprob1}}
where `breves' and `tildes' essentially swap when compared to the revival acceptance probability, indicating the fact that the current and proposed samples have swapped. For completeness, 
\cm{
\breve Z_{k\tilde\iota} 
&= 1 + \!\!\!\!\sum_{\{i'\mid \tilde\Omega_{ki'}=1\}} \frac{p(\breve{\bm a}_{1:k}^{i'\tilde\iota}, \breve{\bm\zeta}_{1:k}^{i'\tilde\iota}, \breve\eta_{1:k}^{i'\tilde\iota}\mid \bm y_{1:k})}{p(\tilde{\bm a}_{1:k}, \tilde{\bm\zeta}_{1:k}, \tilde\eta_{1:k}\mid \bm y_{1:k})},
}{\label{eq:splitnormconstant}}
where we let $\{\breve{\bm a}_{1:k}^{i'\tilde\iota},\breve{\bm\zeta}_{1:k}^{i'\tilde\iota},\breve\eta_{1:k}^{i'\tilde\iota}\}$ be the sample which would result from attaching $\tilde\iota$ to $i'$ were object $i$ split at $\tilde\kappa$. Equivalently, it is the \textit{current} sample, edited to delete track $i$ at $\tilde\kappa$, revive $i'$ and reassociate any data from track $i$ at step $k$ to track $i'$. In context, the probability ratio in which this sample appears (indirectly) represents the probability, were the opportunity present, of exactly reversing the effect of this split with a revival, as opposed to doing a different revival with the $\tilde\iota$. Figure \ref{fig:splitsampletypes} shows a toy example of the different sample types and how they relate to one another.

Upon accepting a revival or split, any relevant quantities (e.g., $\bm m_{kic}$) must be deduced from the accepted sample. Additionally, as the revival and splitting kernels are run over every (valid) track, and these affect the indexing of the various tracks, care should be taken as to the ordering of operations. As an example of requiring an index shift, if tracks $i=2,3$ are initialised at $k$, and $i=1$ gets revived with track 2, then track 3 should now be indexed with $i=2$. 

Firstly, initialise a revival counter $R_{kN_{k-1}}:=0$ and revival record $\bm r_{kN_{k-1}}:=\emptyset$. Then, the revival kernel is to be run on each new track $\iota=N_{k-1}+1,\dots,\breve N_k$ in ascending order. If $\tilde\imath\sim q_r(i\mid\iota)$ is accepted (with $\tilde\imath\ne0$), then the revival is acknowledged with $R_{k\iota} = R_{k,\iota-1} + 1$ and $\bm r_{k\iota} = \bm r_{k,\iota-1} \cup \{\iota\}$. Then, the status and validity variables of $\tilde\imath$ are updated: $\zeta_{\breve\kappa_{\tilde\imath}:k,\tilde\imath}=\bm 1$ and $\Omega_{\breve\kappa_{\tilde\imath}:k,\tilde\imath} = \bm 0$. The observation rate hyperparameters, $\alpha_{\breve\kappa_{\tilde\imath}:k,\tilde\imath}$ and $\beta_{\breve\kappa_{\tilde\imath}:k,\tilde\imath}$, are found by repeated use of \eqref{eq:updategamma3}. Next, the state parameters $\{\bm m_{k\tilde\imath c}, V_{k\tilde\imath c}\}_{c=1}^{N_c}$ and the class probabilities, $\bm\pi_{k\tilde\imath}$, are found, using \eqref{eq:trackpred1}, \eqref{eq:classupdate1}, \eqref{eq:updatedpi1} and \eqref{eq:updatedpi2}, as required. Otherwise, if no change is accepted, then $R_{k\iota}=R_{k,\iota-1}$, $\bm r_{k\iota}=\bm r_{k,\iota-1}$, and track $\iota$ in the current sample becomes track $\iota-R_{k\iota}$ in the final sample.

Having done all of the revivals, initialise $S_{k0}:=\breve N_k-R_{k\breve N_k}$, and run the splitting kernel on all pre-existing tracks $i=1,\dots,N_{k-1}$ such that $\breve\Omega_{ki}=1$ but with $i\notin\bm r_{k\breve N_k}$. If splitting $i$ at $\tilde\kappa$, is accepted, then this is acknowledged with $S_{ki}=S_{k,i-1}+1$. Track $i$ in the final sample becomes the first $\tilde\kappa - \check\kappa_i + 1$ steps of track $i$ in the current sample, and a new track with index $\iota = S_{ki}$ is initialised in the usual way for the final sample. Otherwise, $S_{ki} = S_{k,i-1}$, and track $i$ in the final sample is copied exactly from track $i$ in the current sample.

Finally, we have that $N_k = S_{kN_{k-1}}$, and so $\eta_k = N_k - N_{k-1}$, which allows $\varepsilon_k$ and $\xi_k$ to be updated in the final sample with \eqref{eq:updategamma2}. The final sample associations, $\bm a_k$, can be found with \eqref{eq:proposedsamp2}, if required.

Algorithm \ref{alg:revivalkernelshort} condenses the full revival / split MCMC kernel procedures. To obtain GaPP-ReaCtion, perform a modified Algorithm \ref{alg:gappclass}, recording revival validities $\breve\Omega_{ki}^j$ alongside track deletion. Then, after normalising the particle weights, Algorithm \ref{alg:revivalkernelshort} is done for each particle.

\begin{algorithm}
    \caption{Revival / Splitting at $k$}
    \begin{algorithmic}
        \Require data $\bm y_k$, all quantities for current sample at $k$
        \For{new tracks in current sample}
            \State sample revival with \eqref{eq:qr2}
            \If{change accepted with \eqref{eq:revivalaccprob1}}
                \State revive track and add to final sample
            \Else
                \State add the new track to final sample
            \EndIf
        \EndFor
        \For{valid pre-existing tracks in current sample}
            \State sample split with \eqref{eq:qs}
            \If{accepted with \eqref{eq:splitaccprob1}}
                \State add split track and start of old track to final sample
            \Else
                \State add the old track to final sample
            \EndIf
        \EndFor
        \State deduce number of tracks and births in final sample
        \State update birth rate parameters with \eqref{eq:updategamma2}
        \Ensure all quantities for final sample at $k$
    \end{algorithmic}
    \label{alg:revivalkernelshort}
\end{algorithm}

\section{Experiment}
\label{sec:exp}

\subsection{Performance Metrics}
\label{ss:metrics}

To benchmark performance of these methods, the metrics are here outlined. Firstly, metrics for tracking performance are considered. Typically, one may use some (or all) of the Single Integrated Air Picture (SIAP) metrics \cite{votruba2001}, along with 
the Generalised Optimal Sub-Pattern Assignment (GOSPA) metric \cite{rahmathullah2017generalized}, with $\alpha_G=2$, as strongly suggested by the authors. However, due to the fact that inference for GaPP-Class and GaPP-ReaCtion is based upon a particle filter, and object locations within each of these are given by a distribution, none of these are directly applicable. For GOSPA, point estimates for objects are first found, taking a weighted average of those in each particle and applying an existence threshold, and then GOSPA is calculated on the resulting scene estimate.
For both experiments, the specification of GOSPA chosen is with parameters $(p_G,c_G,\alpha_G)=(2,10,2)$, where $c_G$ is the GOSPA `cut-off' distance, $p_G$ is the GOSPA `order', and $\alpha_G$ is the GOSPA `normalisation' constant.

The SIAP quantities we choose are:
\begin{itemize}
    \item Continuity ($C\in[0,1]$): the proportion of ground truth flight time for which it is associated to an inferred track;
    \item Ambiguity ($A\in[1,\infty)$): the average number of associated tracks per \textit{associated} ground truth;
    \item Spuriousness ($S\in[0,1]$): the average proportion of track flight time for which the track is unassociated to any truth;
    \item Positional Accuracy ($P\in[0,\infty)$): the average root mean squared error (RMSE) over total track flight time;
    \item Rate of Track Breaks ($R\in[0,1)$): the average number of track breaks (i.e., track ID changes) over total associated ground truth flight time.
\end{itemize}
It should be noted that, in practice, we will often instead use m$R$ ($\text{milli}-R = 1000\times R$), and that $R$ is the only metric not computed in real-time. We use a `weighted' version of these SIAP metrics to account for the probabilistic nature of the results of the presented methods. Using $A$ as an example, let $\aleph_k$ be the number of active tracks associated to an active true trajectory at step $k$, and $\tilde\aleph_k$ be the number of active true trajectories with associated tracks. Then, $A = \frac{\sum_k \aleph_k}{\sum_k \tilde\aleph_k}$.
The `weighted' equivalent we use is $A_w = \frac{\sum_k \sum_j w_k^j \aleph_k^j}{\sum_k \sum_j w_k^j \tilde\aleph_k^j}$,
where $\aleph_k^j$ and $\tilde\aleph_k^j$ are the same as above, but for particle $j$.

The average real-time RMSE is used to assess the hyperparameter learning and classification. In the experiments presented here, only two classes of object are used (with class labels $\mathcal C_\iota\in\{0,1\}$), which will allow for a simplified metric, discussed later. Meanwhile, let $\gamma^*$ be the true value of $\gamma$ (and similarly for other `starred' parameters). Then, the mean squared error of $\gamma$ at time step $k$ is found from its posteriors in each particle in \eqref{eq:updategamma1}, as
\cm{
\mathbb E(&(\gamma-\gamma^*)^2\mid \bm y_{1:k})\\
&\approx \sum_{j=1}^J w_k^j \left( \frac{\varepsilon_k^j(\varepsilon_k^j+1)}{(\xi_k^j)^2} - 2\gamma^*\frac{\varepsilon_k^j}{\xi_k^j} + (\gamma^*)^2 \right).
}{\label{eq:msegamma}}
Then, the average RMSE (aRMSE) is just the square root of the result of \eqref{eq:msegamma}, averaged over all $K$ time steps,
\cm{
\text{aRMSE}(\gamma) = \frac1K\sum_{k=1}^K \sqrt{\mathbb E\left((\gamma-\gamma^*)^2\mid\bm y_{1:k}\right)}.
}{\label{eq:armsegamma}}
Similar results are used for $\mu_0$ and $s^2$, from their Gamma and Inverse Gamma distributions, respectively, whilst the aRMSE for the local hyperparameters is additionally averaged over truths, weighted by the number of time steps for which they have associated tracks. We denote this combined aRMSE of all detection rates, $\mu_i$, as $\text{aRMSE}(\mu_{>0})$, and that of all classification variables, $\mathcal C_i$, as $\text{aRMSE}(\mathcal C)$.
Thus, we have established tracking metrics to assess and compare the proposed methods, and some further metrics to assess hyperparameter learning.

\subsection{Alternative Models for Comparison}
\label{ss:rivals}

In order to assess the capability of the proposed methods compared to the state of the art, we run a collection of other multi-object trackers on the same data, with comparable design choices. In particular, the other trackers are
\begin{itemize}
    \item DiGiT \cite{goodyer2023flexible}, using a GP-type dynamical model and a Dirichlet Process-based inference model,
    \item a Message Passing (MP) approach, e.g, \cite{meyer2018message}, with implementation with \cite{meyer2021}, with an Interacting Multiple Model (IMM) \cite{blom1988interacting} for the dynamics,
    \item the same MP-based method, but with a Constant Velocity (CV) dynamical model,
    \item a Gaussian Mixture (GM) Probability Hypothesis Density (PHD) filter \cite{panta2004,vo2006} with CV dynamics, implemented in Stone Soup \cite{barr2022stone}, a Python library, and 
    \item a Global Nearest Neighbour (GNN) with CV dynamics, within Stone Soup, serving as a simple baseline method.
\end{itemize}
State of the art methods assuming at most one observation per time step (e.g., TPMBM) are not used, as these data sets severely violate this assumption. Where possible, computational complexity is roughly tailored to make meaningful comparisons. Specifically, the number of particles chosen for the proposed GaPP-based methods and the message passing ones are selected such that the run time over all of the synthetic data sets in Subsection \ref{ss:syntheticexp} are comparable. These particle numbers are then approximately retained (up to scaling) for the experiment on the real drone surveillance data, in Subsection \ref{ss:realexp}. The different motion models chosen for the IMM are a CV model, a Constant Acceleration (CA) model, and a Constant Turn (CT) model.

\subsection{Synthetic Data Experiment}
\label{ss:syntheticexp}

\begin{figure}[thbp]
    \centering
    \includegraphics[width=1\linewidth]{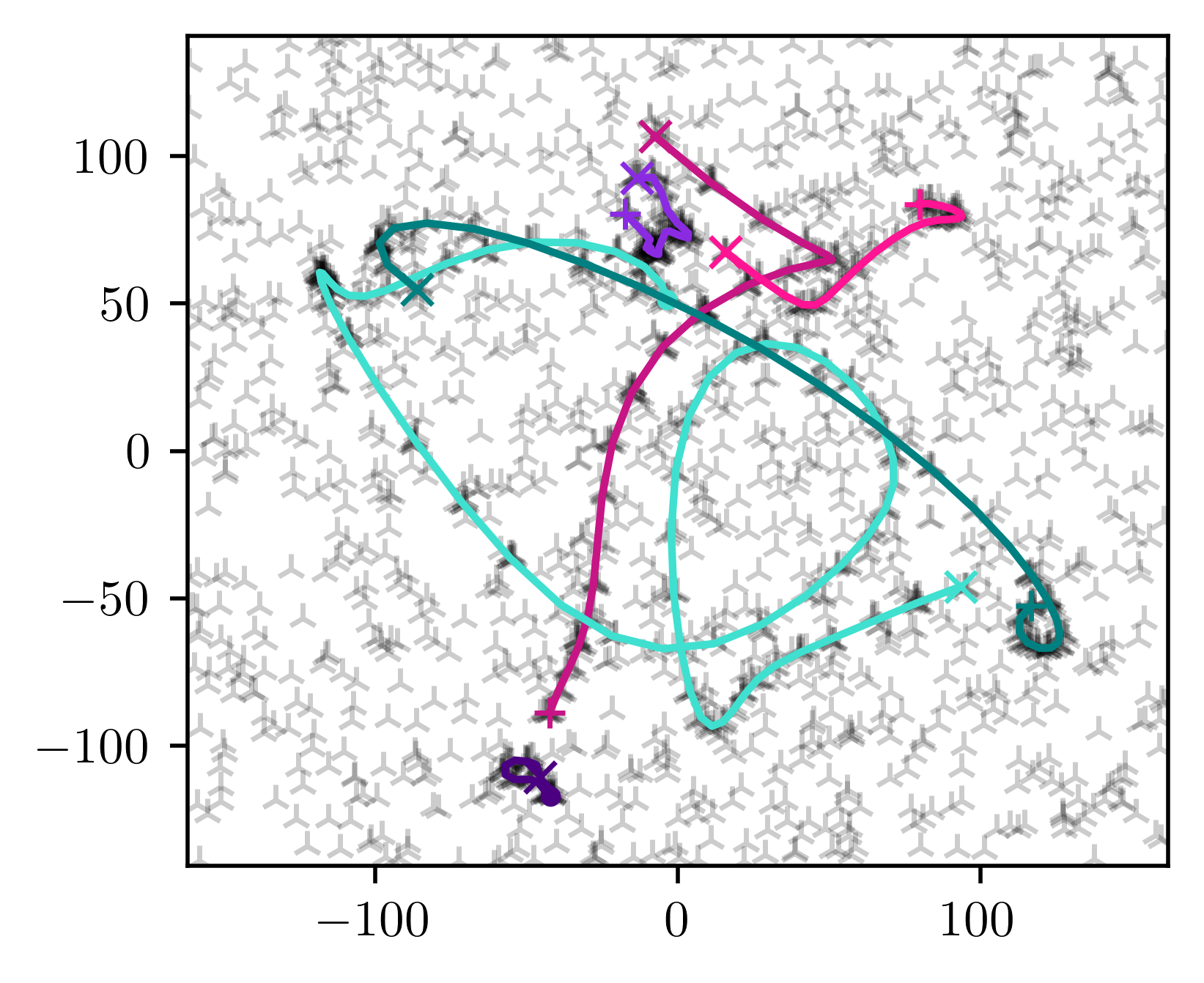}
    \caption{Example synthetic data set. Trajectories begin with `$\times$', and end with `$+$' if it was deleted, and `$\bullet$' if not, whilst observations are three-pointed stars.}
    \label{fig:synthset6}
\end{figure}

One hundred synthetic data sets, each consisting of 100 equal time steps of size 1, have their global parameters sampled uniformly from the following intervals.
\cm{
& \mu_0 \sim U(10,15), && \gamma\sim U(0.02,0.1), && s^2\sim U(0.5,2)
}{\label{eq:synthglobalparsamp}}
Objects are initialised with local hyperparameters independently sampled from
\cm{
& \mu_i\sim U(3,6), && \mathcal C_i\sim \text{Bernoulli}(0.5), ~\forall\ i,
}{\label{eq:synthlocalparsamp}}
where the iSE hyperparameters for each of the two classes are 
\cm{
&(\sigma_0^2,\ell_0) = (100,4), && (\sigma_1^2,\ell_1) = (10,1).
}{\label{eq:synthlocalclasspars}}
The sliding window length is $d=10$ for both classes. The survival probability for all objects is $\psi = 0.98$ after it has been active for $d$ steps, to reduce the number of very short trajectories. An additional object is started at the first time step, to reduce the number of time steps for which there are no objects present, and no objects are born in the final $d$ steps, again, to reduce very short trajectories. Object trajectories are generated (and translated sensibly for a less sparse scene), after which all observations and clutter are sampled exactly by the NHPP described in Subsection \ref{ss:obsmodel}. An example data set is shown in Figure \ref{fig:synthset6}.

The initial hyperparameters of the prior distributions for the GaPP methods were
\cm{
 (\alpha'_+,\beta'_+) &= (9,0.75), & (\alpha_+,\beta_+) &= (4,1),\\
(\varepsilon_+,\xi_+) &= (0.05,1), & (\varphi_+,b_+) &= (3,2),
}{\label{eq:inithyperpars}}
leading to prior means for parameters being
\cm{
& \mathbb E(\mu_0) = 12, && \mathbb E(\mu_i) = 4, && \mathbb E(\gamma) = 0.05, && \mathbb E(s^2) = 1.
}{\label{eq:priorhyperparmeans}}
The iSE hyperparameters were given (including the window length, $d$), and the prior class probabilities were $\bm\pi_+=\frac12\bm1_2$. The heuristic criteria for deletion were a location standard deviation of more than 50, no observations associated for three or more consecutive time steps, or if ever $\mathbb E(\mu_i\mid \bm a_{1:k}^j,\bm\zeta_{1:k}^j,\eta_{1:k}^j,\bm y_{1:k})<\frac12$. For GaPP-ReaCtion, the maximum number of time steps for revival was $d_\zeta = 3$. Both methods use $J=50$ particles. All methods, except the baseline GNN-CV, use a survival probability of $\psi = 0.98$.

For the other methods which used any of the above parameters (except the number of particles), the same value was given, or, in the case of $\mu_0,\mu_i,\gamma$ or $s^2$, the prior mean was given. Models that used a CV had a driving noise variance of 100. For the IMM, the CV component was the same, the CA model had a driving noise variance of 25, and the CT model had a driving noise variance of 0.5 (radians per time step). The initial model probabilities were $(0.5, 0.25, 0.25)$ for the CV, CA and CT, respectively. The model transition matrix was $\frac16\bm1_{3\times3} + \frac12I_3$, so each particle had a 1/3 chance of switching model, with equal probability of going to each of the other two. For both of the MP methods, the pruning and detection thresholds were set at 0.05, with a minimum track length of 1, and 2 MP iterations. The MP-CV used 3000 particles, whilst the MP-IMM used only 300, to match the computational complexities. For DiGiT, the Dirichlet process concentration parameter \cite{teh2010dirichlet} was set to the prior mean of $\gamma$, 0.05, and the shape parameters of the prior composite Dirichlet process mixture \cite{carevic2016} were set to the prior means of $\mu_0$ and $\mu_i$, 12 and 4. The GP hyperparameters were taken as a weighted average of those of the two classes, specifically in a 4:1 ratio resulting in $(\sigma^2,\ell)=(82,3.2)$. The sliding window for observation counts was taken as the same as $d_\zeta=3$ from GaPP-ReaCtion. The number of particles was $J=100$, which gave greater computational cost, but fewer particles led to too poor and unreliable performance. 

The GM-PHD used a pruning threshold of $10^{-6}$, a detection threshold of 0.9, a merge threshold of 100, and a maximum number of PHD components of 100. The GNN-CV used a maximum observation Mahalanobis distance of 3, a missed detection limit of 2, and a positional covariance limit of 100, and a minimum number of points to construct a track as 5.

The tracking results (not including hyperparameter learning or classification) are shown in Table \ref{tab:syntheticresults}. `Time' is averaged over data sets, and given in seconds.
\begin{table}[ht]
    \centering
    \begin{tabular}{PcRcRcRc}
        \rowcolor{palepurple} \textcolor{white}{\textbf{Metrics}} & \textcolor{white}{$C$} & \textcolor{white}{$A$} & \textcolor{white}{$S$} & \textcolor{white}{$P$} & \textcolor{white}{m$R$} & \textcolor{white}{GOSPA} & \textcolor{white}{Time} \\
        \hline
        \textcolor{white}{GaPP-} &&&&&&& \\
        \textcolor{white}{ReaCtion} & \multirow{2}{*}[3mm]{\textbf{0.99}} & \multirow{2}{*}[3mm]{\textbf{1.00}} & \multirow{2}{*}[3mm]{0.04} & \multirow{2}{*}[3mm]{\textbf{0.76}} & \multirow{2}{*}[3mm]{\textbf{2.16}} & \multirow{2}{*}[3mm]{1.99} & \multirow{2}{*}[3mm]{10.7} \\
        \hline
        \textcolor{white}{GaPP-Class} & \textbf{0.99} & \textbf{1.00} & 0.04 & \textbf{0.76} & 2.54 & \textbf{1.95} & 7.99 \\
        \hline
        \textcolor{white}{DiGiT} & 0.85 & \textbf{1.00} & 0.13 & 0.82 & 37.8 & 4.05 & 24.5 \\
        \hline
        \textcolor{white}{MP-IMM} & 0.98 & 1.02 & 0.11 & 1.18 & 52.7 & 3.65 & 11.0 \\
        \hline
        \textcolor{white}{MP-CV} & 0.98 & 1.01 & 0.06 & 0.82 & 14.3 & 2.33 & 9.71 \\
        \hline
        \textcolor{white}{GM-PHD} & 0.92 & \textbf{1.00} & \textbf{0.02} & 0.90 & 11.0 & 2.65 & 10.9 \\
        \hline
        \textcolor{white}{GNN-CV} & 0.97 & 2.86 & 0.08 & 1.56 & 30.2 & 15.0 & \textbf{1.01} \\
        \hline
        \rowcolor{palepurple} \textcolor{white}{Optimal} & \textcolor{white}{1} & \textcolor{white}{1} & \textcolor{white}{0} & \textcolor{white}{0} & \textcolor{white}{0} & \textcolor{white}{0} & \textcolor{white}{0} \\
    \end{tabular}
    \caption{Tracking results averaged over time steps / data sets for synthetic data experiments for all methods, where \textbf{bold} values indicate the best performance.}
    \label{tab:syntheticresults}
\end{table}

The baseline GNN-CV was fastest, as expected due to its simplicity, but the results are poor, due to the high $A$ and GOSPA values showing that it produced far too many tracks for each truth, where all other methods were able to keep $A$ close to 1, and GOSPA lower. As a result, the GOSPA value is incredibly high compared to the other methods. DiGiT also stands out for having low $C$, meaning it failed to track the objects as successfully as the other methods. A mitigating factor is that, compared to the GaPP- and MP- approaches, and the PHD to a lesser extent, the Dirichlet Process as the measurement model is a further departure from the model assumptions compared to the truth. DiGiT is also the most expensive method.

The MP-IMM seems slightly worse than the best performing methods across all metrics, but with significantly more track breaks, so a higher m$R$. The GM-PHD is accurate and produces the fewest false tracks, but, of the remaining methods, the $C$ value is much lower, preventing a better GOSPA score. The GaPP methods and the MP-CV are the best performing methods, with MP-CV marginally outperformed across all metrics. The GaPP methods perform almost identically, with the exception of the track breaks, which are reduced in GaPP-ReaCtion for a small computational cost.

\begin{figure*}[tbhp]
    \centering
    \begin{subfigure}{0.31\linewidth}
        \includegraphics[width=\linewidth]{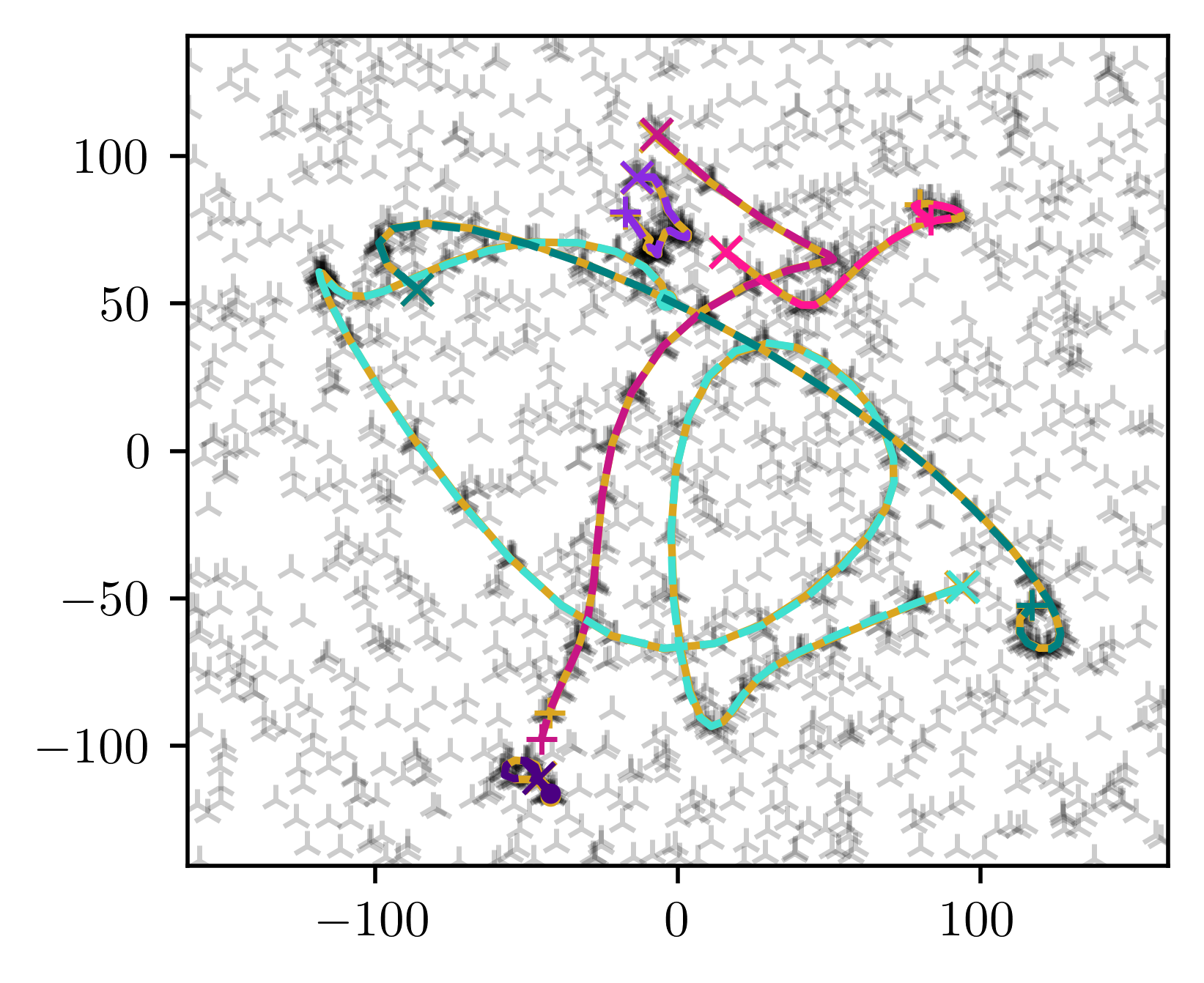}
        \caption{GaPP-ReaCtion}
        \label{subfig:synthreaction}
    \end{subfigure}
    \begin{subfigure}{0.31\linewidth}
        \includegraphics[width=\linewidth]{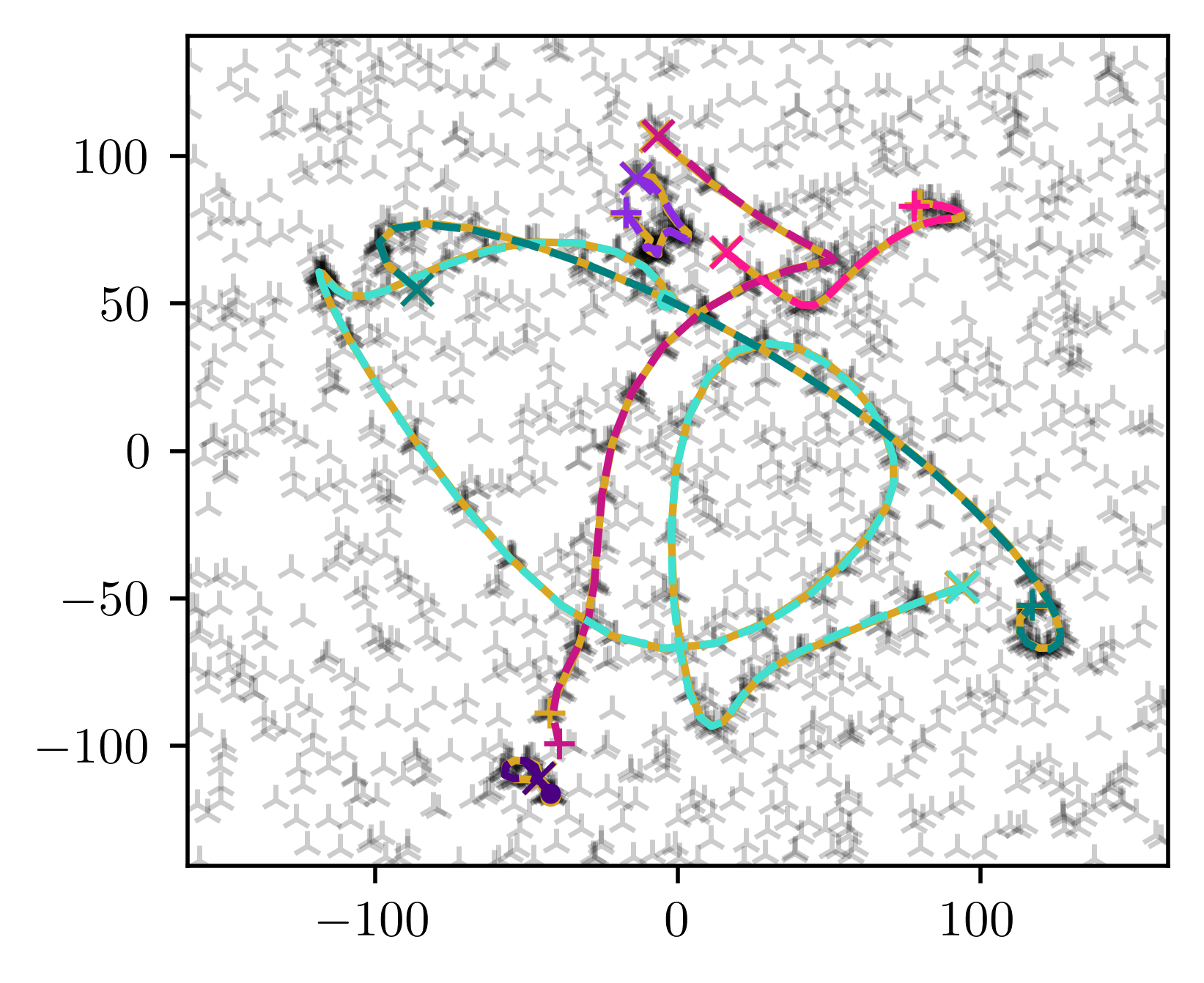}
        \caption{GaPP-Class}
        \label{subfig:synthclass}
    \end{subfigure}
    \begin{subfigure}{0.31\linewidth}
        \includegraphics[width=\linewidth]{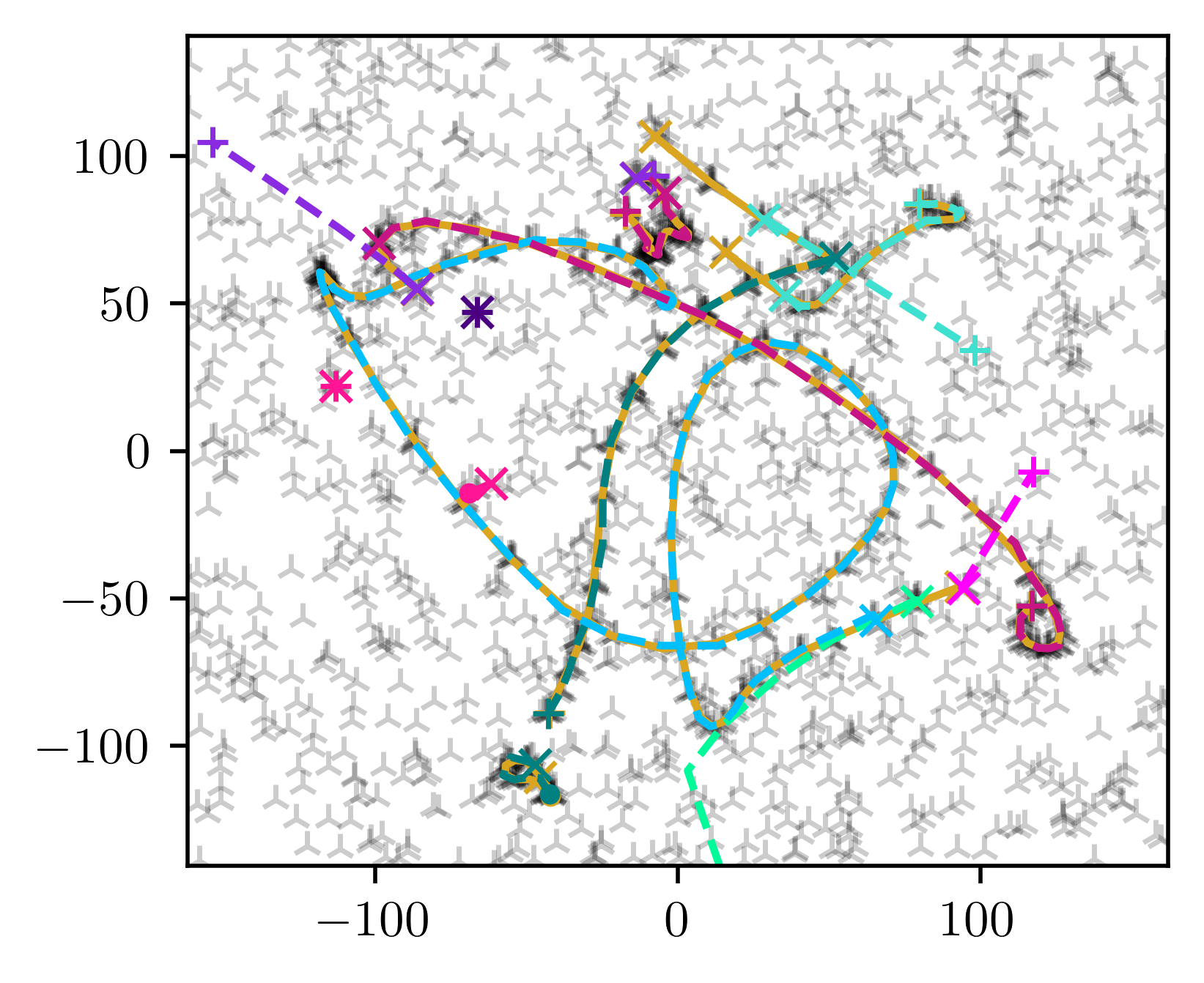}
        \caption{DiGiT}
        \label{subfig:synthdigit}
    \end{subfigure}
    \begin{subfigure}{0.31\linewidth}
        \includegraphics[width=\linewidth]{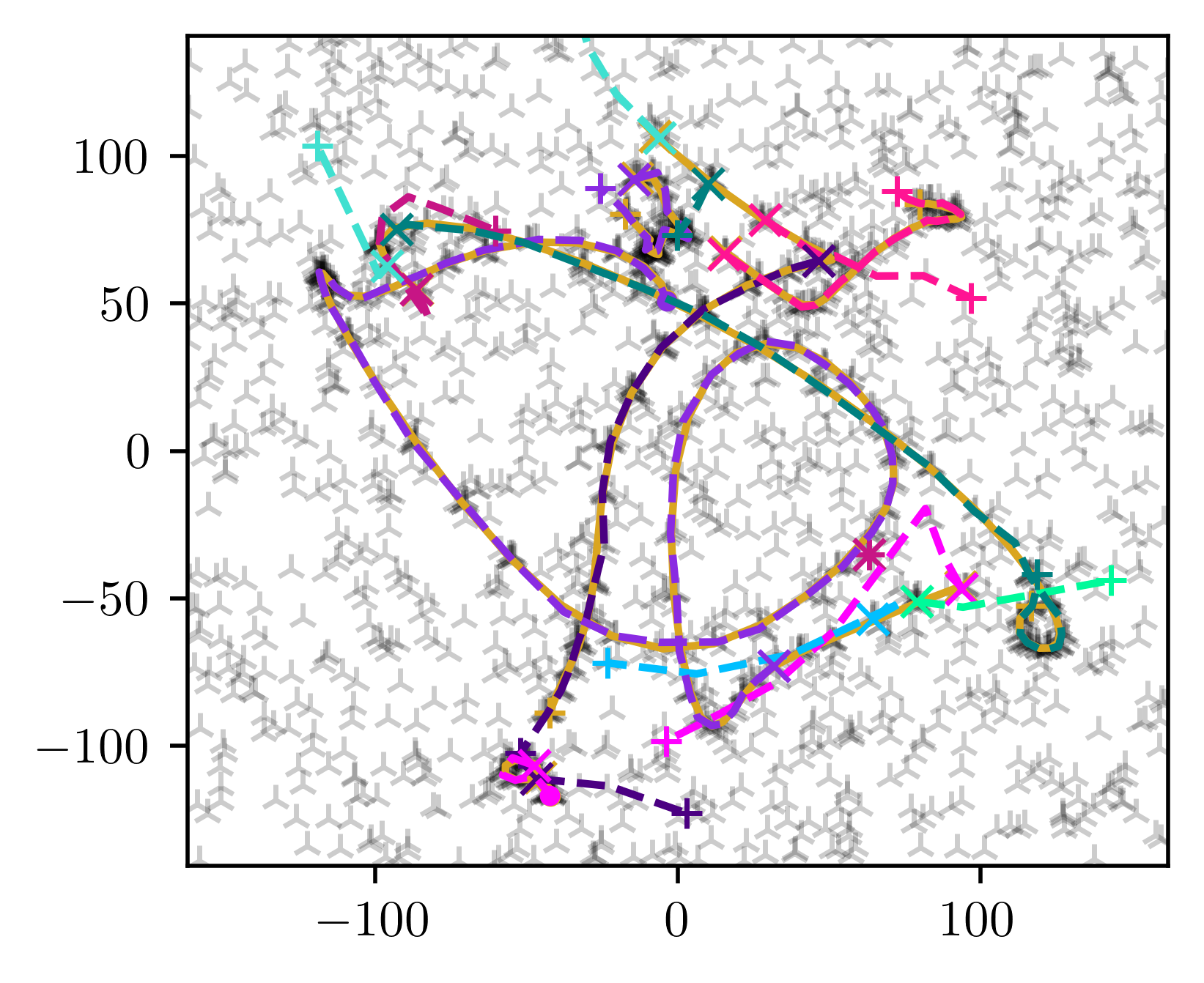}
        \caption{MP-IMM}
        \label{subfig:synthmpimm}
    \end{subfigure}
    \begin{subfigure}{0.31\linewidth}
        \includegraphics[width=\linewidth]{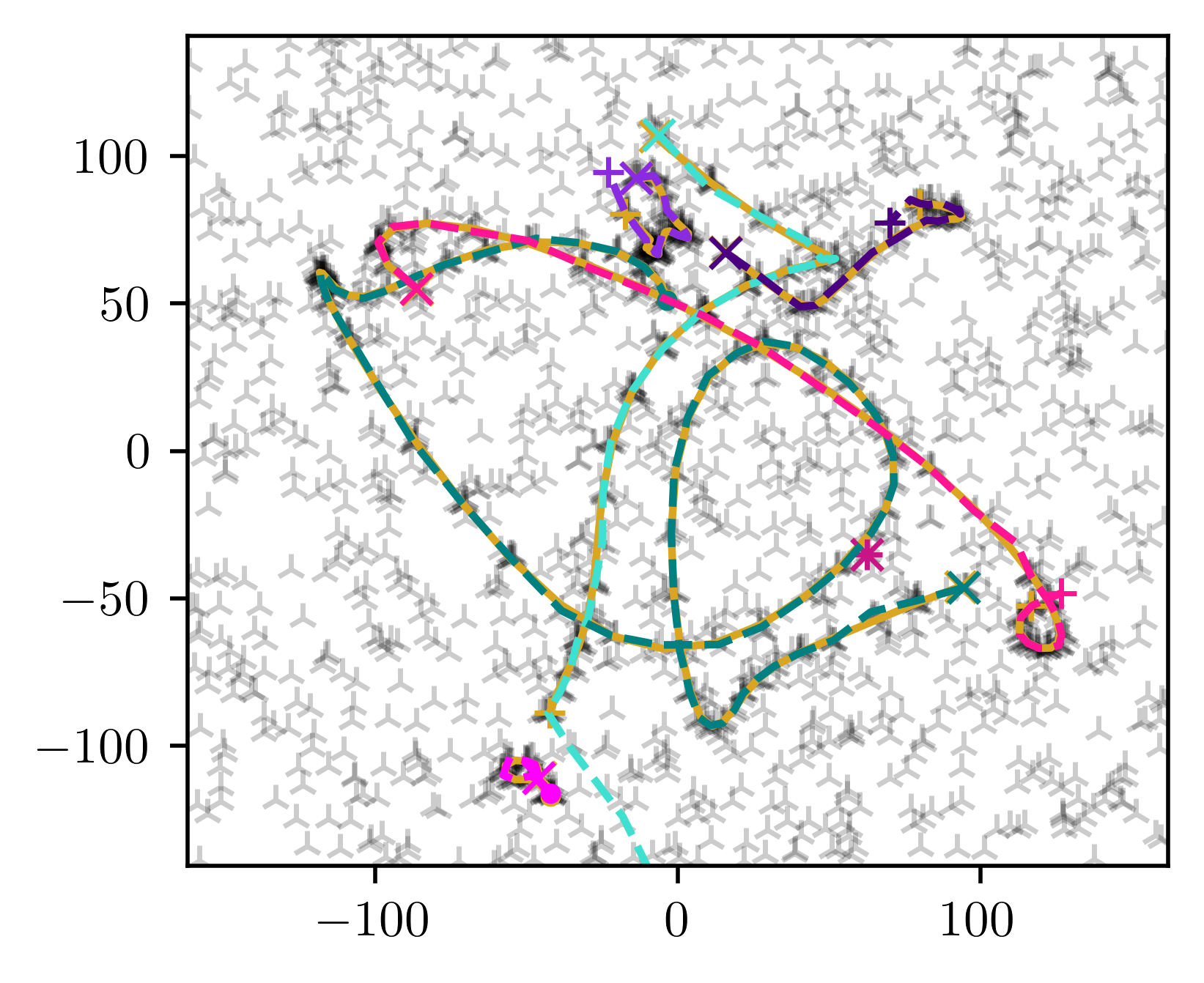}
        \caption{MP-CV}
        \label{subfig:synthmpcv}
    \end{subfigure}
    \begin{subfigure}{0.31\linewidth}
        \includegraphics[width=\linewidth]{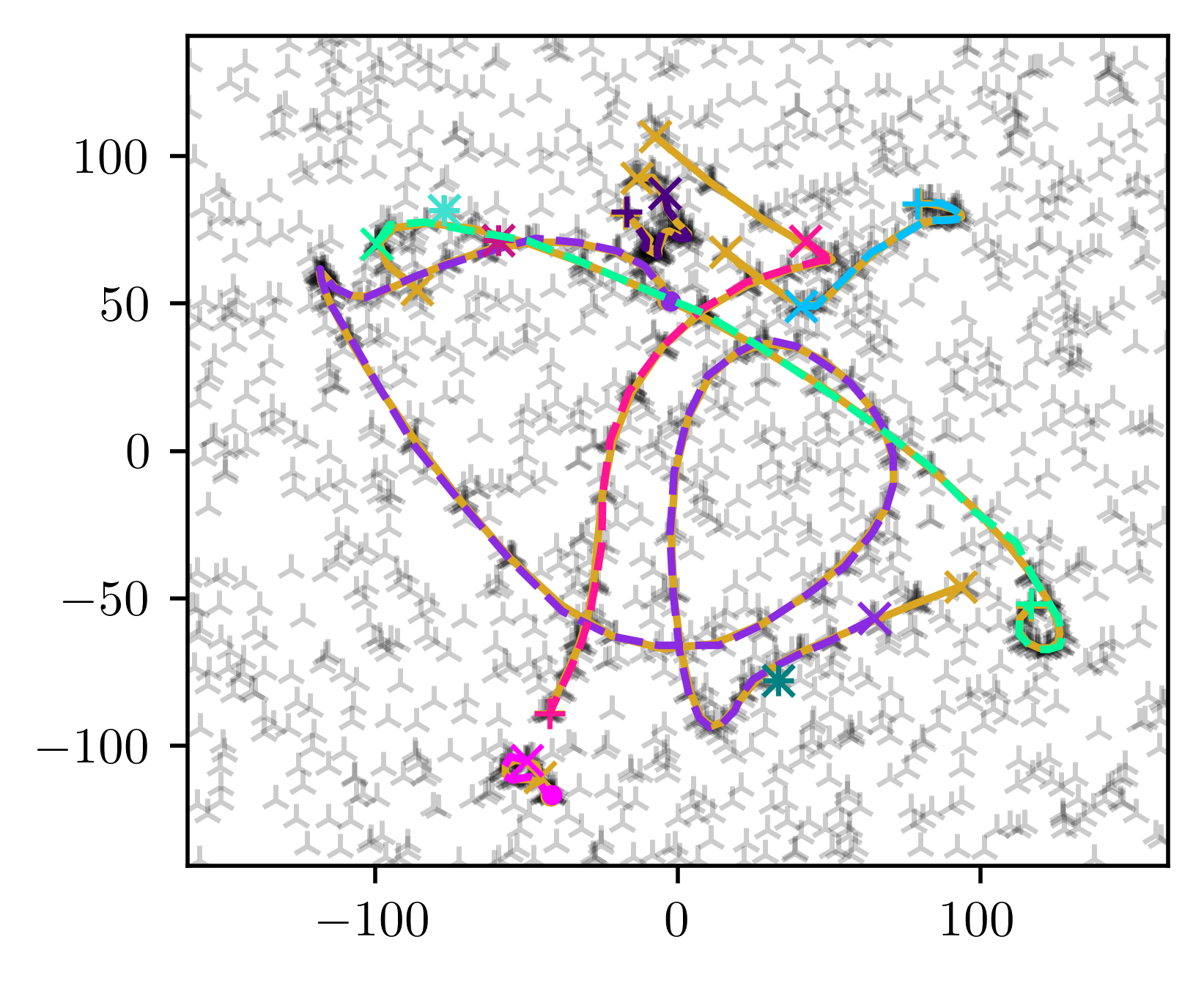}
        \caption{GM-PHD}
        \label{subfig:synthgmphd}
    \end{subfigure}
    \caption{Final track estimates from a synthetic data set; `$\times$' is track start, `$\bullet$'/`$+$' is a surviving/deleted track's end. Ground truth object trajectories are gold, and data are grey stars.}
    \label{fig:syntheticresults6}
\end{figure*}

\begin{table}[ht]
    \centering
    \begin{tabular}{PcRcRcR}
        \rowcolor{palepurple} \textcolor{white}{\textbf{(a)RMSE}(...)} & \textcolor{white}{$s^2$} & \textcolor{white}{$\gamma$} & \textcolor{white}{$\mu_0$} & \textcolor{white}{$\mu_{>0}$} & \textcolor{white}{$\mathcal C$} \\
        \hline
        \textcolor{white}{GaPP-ReaCtion} & \textbf{1.52} & \textbf{0.08} & \textbf{0.86} & \textbf{0.73} & \textbf{0.08} \\
        \hline
        \textcolor{white}{GaPP-Class} & 1.58 & 0.09 & \textbf{0.86} & 0.74 & \textbf{0.08} \\
    \end{tabular}
    \caption{Hyperparameter learning and classification results averaged over time steps and synthetic datasets for GaPP-ReaCtion and GaPP-Class; \textbf{bold} indicate the best performance.}
    \label{tab:synthetichyperresults}
\end{table}

Points of note include the fact that the MP-CV comprehensively outperformed the MP-IMM. This is because the additional complexity of the IMM reduced the number of particles compared to its CV counterpart (to retain similar computational complexity). This highlights the importance of the inference scheme and measurement model even for fast-manoeuvring objects compared to flexible dynamics. The flexibility in the CV is enabled by simply choosing the driving noise variance to be very large. It is for this reason that the IMM was not used for the other methods using a CV. Aside from this, the ability of the GaPP methods to use multiple object classes seems to permit fewer track breaks than the rest when these classes are present, improving robustness. The revival kernel contributes further to this. They are also similarly fast compared with the other methods (except the baseline) despite expending additional resources for hyperparameter learning.

The final tracks resulting from inference on the aforeshown exemplar data set are shown in Figure \ref{fig:syntheticresults6}, in which we see almost perfect performances from GaPP-ReaCtion (fig. \ref{subfig:synthreaction}) and GaPP-Class (fig. \ref{subfig:synthclass}), and a good performance by MP-CV (fig. \ref{subfig:synthmpcv}). The GNN-CV is not shown in Figure \ref{fig:syntheticresults6} since it performs very poorly compared to the other methods.


Hyperparameter learning and classification results for the two GaPP methods are found in Table \ref{tab:synthetichyperresults}, wherein the performance for all hyperparameters are almost identical for both methods, albeit slightly in favour of GaPP-ReaCtion, possibly indicating that it can represent the true posterior more accurately for a given number of particles. It is especially interesting that, in this experiment, the classification accuracy is equal for both. In this case, it may be because the iSE hyperparameters for each class yield such different tracks on average, that classification was often trivial. 

Another significant result is that the learning of $s^2$ is difficult, implied by its much larger RMSE for both GaPP methods in Table \ref{tab:synthetichyperresults}, even larger than the RMSE for $\mu_0$, which was nominally an order of magnitude greater. This could be due to the fact that, if the model thinks that $s^2$ is very high, then it can very well explain the data whilst reducing track initialisations and deletions, reducing extreme motion of objects, and increasing observation likelihood for would-be clutter, since more data are attributable to objects. All of those aspects of a tracking scenario can be considered unlikely events. At the very least, there may be stable equilibria in the hyperparameter space which have high $s^2$ values, which should be prevented by using a more informative prior. It should be noted that the performance here on $s^2$ is relatively unstable, and so minor changes to \textit{any} hyperparameters can significantly change the learnt distribution over $s^2$, without changing the tracking inference or the other learnt parameter distributions. This remains an area of interest for future work.

\subsection{Real-World Radar Surveillance Experiments with Drones}
\label{ss:realexp}

\begin{figure}[thbp]
    \centering
    \includegraphics[width=0.7\linewidth]{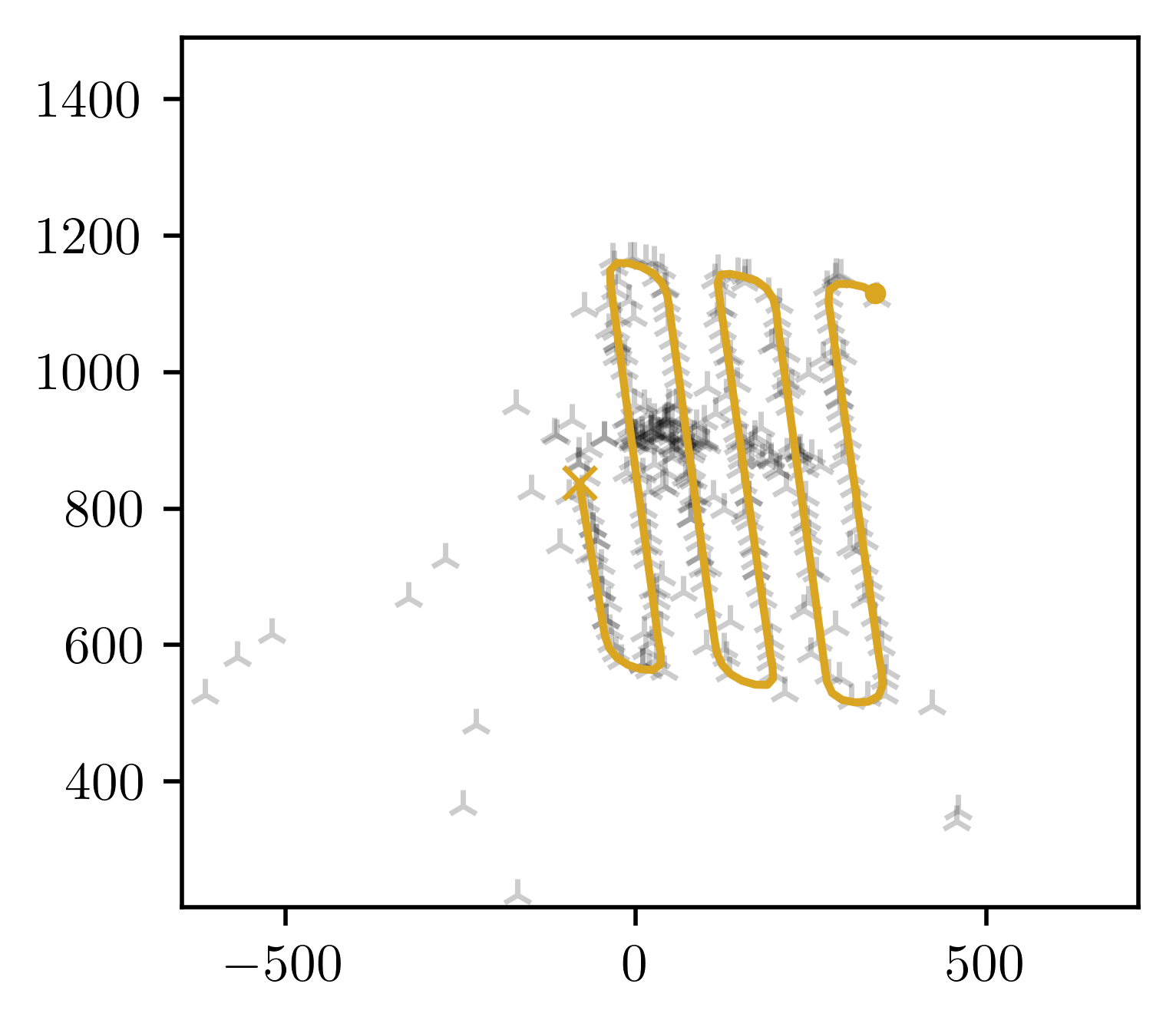}
    \caption{Example real radar measurements data set; `$\times$' and `$\bullet$' are the start and end of the ground truth drone trajectory.
    }
    \label{fig:realset9}
\end{figure}

Here, we use real radar measurements from six live drone trials with five different drone platforms (all class I, sub-5kg, platforms) across five different complex semi-urban environments. This data is from Thales' GameKeeper radar (max.\ range $\approx 7.5$km) for drone surveillance applications, for example for fixed-site protection such as civilian airports. Distance of the small drone targets (e.g., DJI Inspire 2, with weight $\approx 4$kg) from the sensor ranges from 250m to over 3km. For simplicity, we process observations (i.e., radar plots or detections) that only contain 2D coordinates with a (down)sampled update period of 2.22 seconds. The ground truth is available from an on-board navigation system attached to the drone, but not for any other objects in the scene. As a result, only some of our metrics are applicable, namely $C$, $A$, $P$ and $R$. Relative run times should be similar to the first experiment, and the real data sets are highly varied (e.g., number of time steps, number of observations etc.), so average run time is not shown. Technically, $A$ could be compromised if another object (e.g., a bird) is in very close proximity to the drone. This is rare, thus, $A$ is still shown, albeit could be less reliable. Figure \ref{fig:realset9} shows a simple example data set, featuring long periods of constant velocity movement separated by sharper manoeuvres. 

The hyperparameters of the priors of the GaPP-based models were changed to 
\cm{
(\alpha'_+,\beta'_+) &= (10,1), & (\alpha_+,\beta_+) &= (50,10),\\
(\varepsilon_+,\xi_+) &= (0.01,1), & (\varphi_+,b_+) &\approx (1.6,64)\times10^3,
}{\label{eq:realinithyperpars}}
leading to the new prior means being
\cm{
& \mathbb E(\mu_0) = 10, & \mathbb E(\mu_i) = 5, && \mathbb E(\gamma) = 0.01, && \mathbb E(s^2) = 40.
}{\label{eq:realpriorhyperparmeans}}
The measurement errors in the real data were highly variable in size (and Gaussianity), so $(\varphi_+,b_+)$ were chosen to keep $\mathbb E(s^2)$ high, but also the variance low, for the reasons discussed in Subsection \ref{ss:syntheticexp}. The heuristic deletion criteria were retained, but the location standard deviation limit was increased to 1000, all but removing it. Two iSE classes were again assumed, with the parameters being
\cm{
&(\sigma_0^2,\ell_0) = (40,2), && (\sigma_1^2,\ell_1) = (100,8),
}{\label{eq:reallocalclasspars}}
attempting to have a slower, more manoeuvrable class and a faster, less agile one. The number of particles for both GaPPs was doubled from the synthetic experiments, to $J=100$. Similarly, the number of particles for MP-CV was also doubled, this time to 6000. To give the MP-IMM the best possible chance, the number of particles more than tripled, taking it to 1000. Since DiGiT was the most expensive, the number of particles was kept at 100, and the GP parameters chosen were of the more flexible class, so $(40,2)$. Parameters which previously used prior means from the GaPP methods do so again, now from \eqref{eq:reallocalclasspars}. No further changes to the parameters of any method occurred. 

\begin{table}[ht]
    \centering
    \begin{tabular}{PcRcRc}
        \rowcolor{palepurple} \textcolor{white}{\textbf{Metrics}} & \textcolor{white}{$C$} & \textcolor{white}{$A$} & \textcolor{white}{$P$} & \textcolor{white}{m$R$} \\
        \hline
        \textcolor{white}{GaPP-ReaCtion} & 0.86 & 1.07 & 16.1 & \textbf{22.4} \\
        \hline
        \textcolor{white}{GaPP-Class} & 0.86 & 1.07 & 16.1 & 32.5 \\
        \hline
        \textcolor{white}{DiGiT} & \textbf{0.90} & 1.03 & \textbf{15.8} & 37.0 \\
        \hline
        \textcolor{white}{MP-IMM} & 0.62 & 1.12 & 16.8 & 57.2 \\
        \hline
        \textcolor{white}{MP-CV} & 0.65 & 1.10 & 17.6 & 52.0 \\
        \hline
        \textcolor{white}{GM-PHD} & 0.77 & \textbf{1.01} & 16.2 & 32.0 \\
        \hline
        \textcolor{white}{GNN-CV} & 0.82 & 4.85 & 16.0 & 56.6 \\
        \hline
        \rowcolor{palepurple} \textcolor{white}{Optimal} & \textcolor{white}{1} & \textcolor{white}{1} & \textcolor{white}{0} & \textcolor{white}{0}\\
    \end{tabular}
    \caption{Tracking results averaged over time steps for real data experiments for all methods, where \textbf{bold} values indicate the best performance of any model by that metric.}
    \label{tab:realresults}
\end{table}

\begin{figure*}[tbhp]
    \centering
    \begin{subfigure}{0.31\linewidth}
        \includegraphics[width=\linewidth]{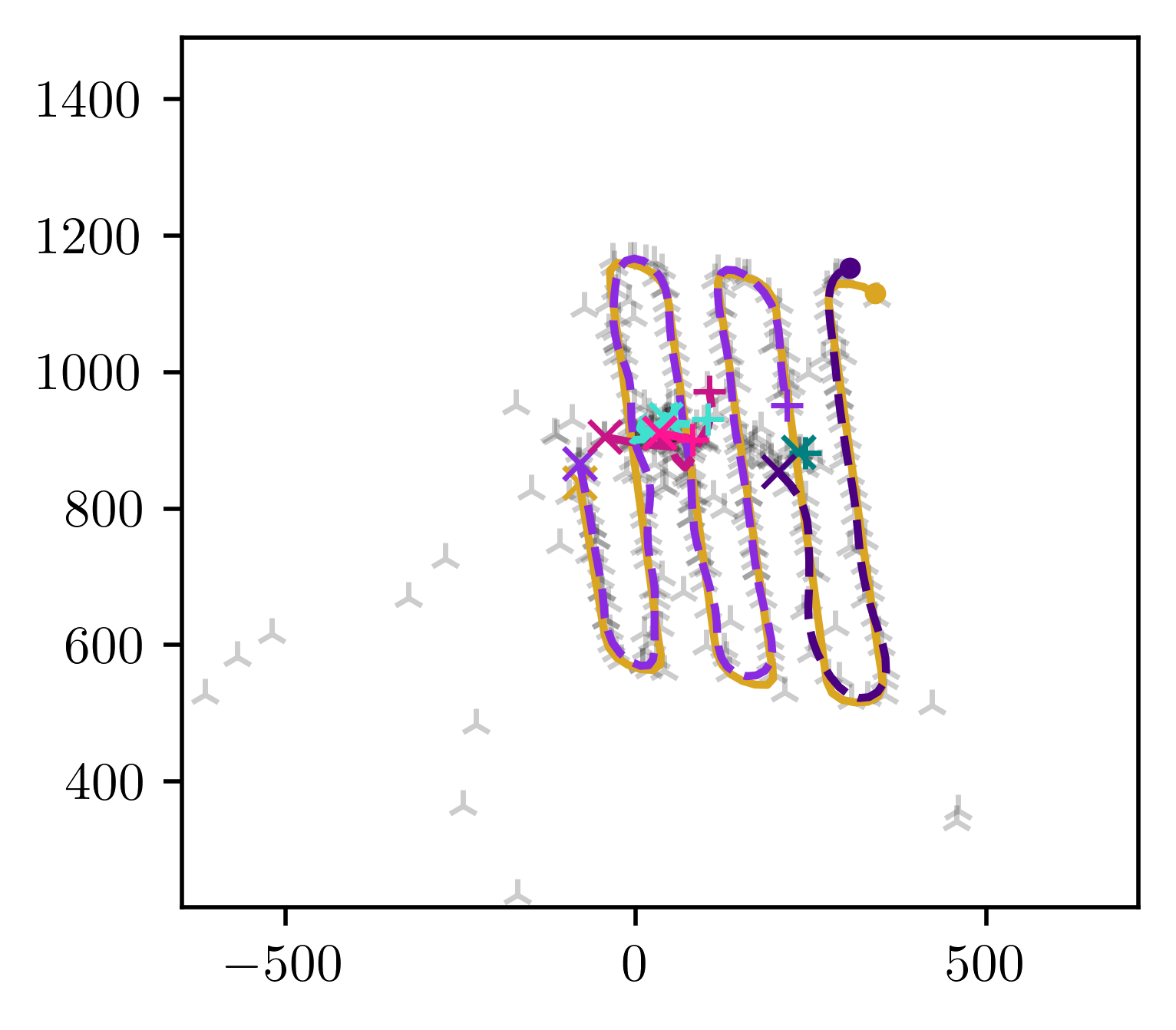}
        \caption{GaPP-ReaCtion}
        \label{subfig:realreaction}
    \end{subfigure}
    \begin{subfigure}{0.31\linewidth}
        \includegraphics[width=\linewidth]{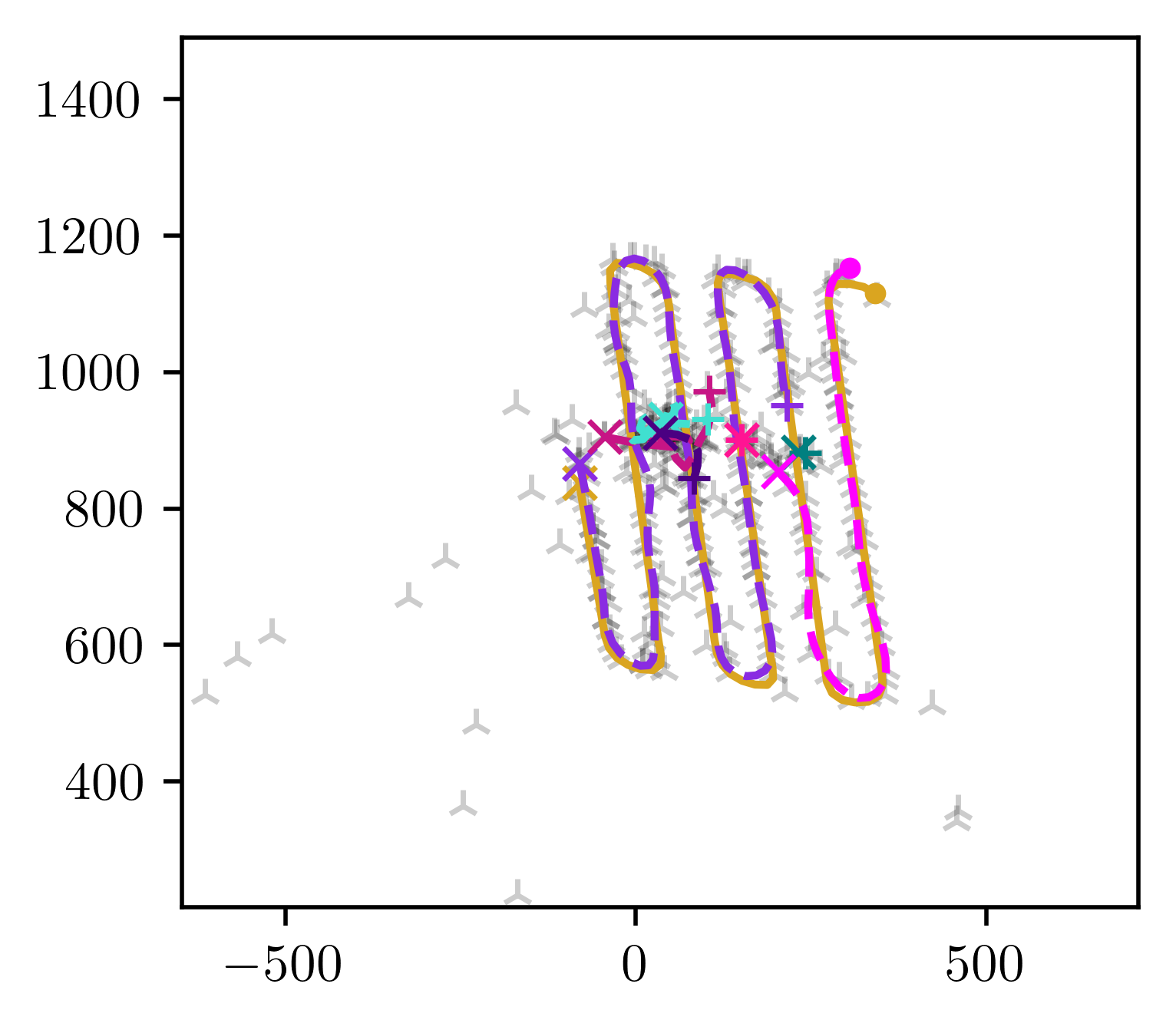}
        \caption{GaPP-Class}
        \label{subfig:realclass}
    \end{subfigure}
    \begin{subfigure}{0.31\linewidth}
        \includegraphics[width=\linewidth]{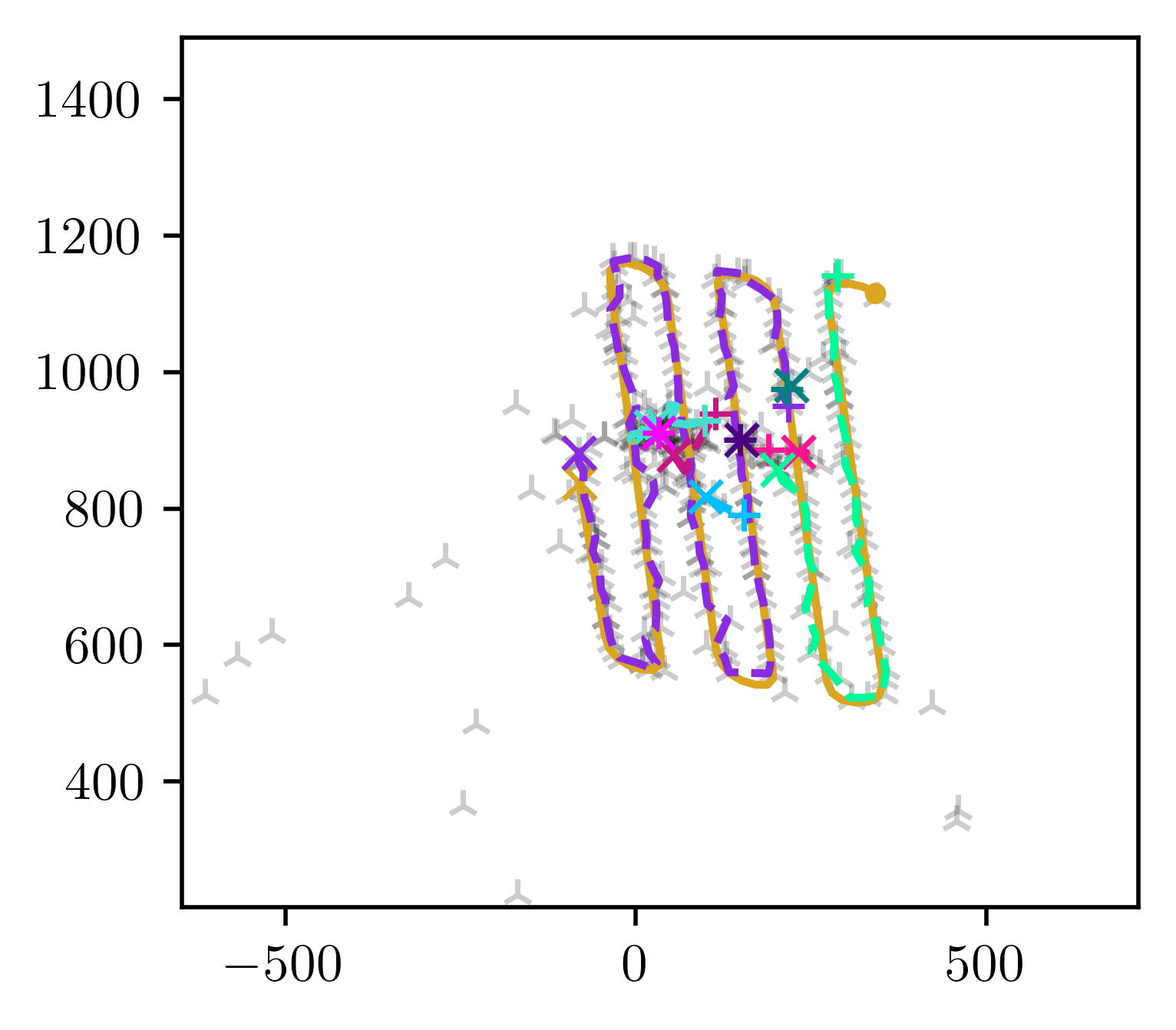}
        \caption{DiGiT}
        \label{subfig:realdigit}
    \end{subfigure}
    \begin{subfigure}{0.31\linewidth}
        \includegraphics[width=\linewidth]{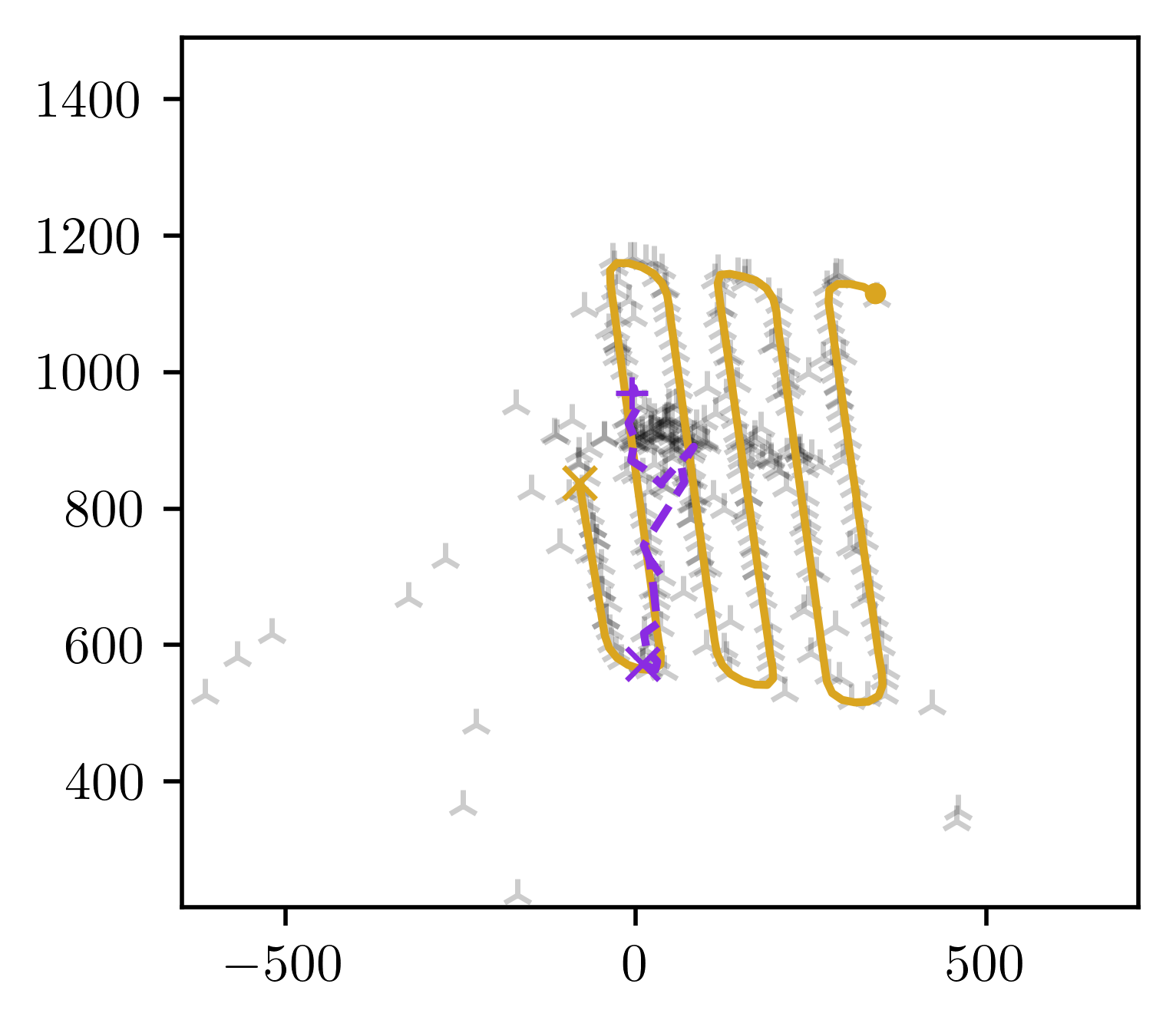}
        \caption{MP-IMM}
        \label{subfig:realmpimm}
    \end{subfigure}
    \begin{subfigure}{0.31\linewidth}
        \includegraphics[width=\linewidth]{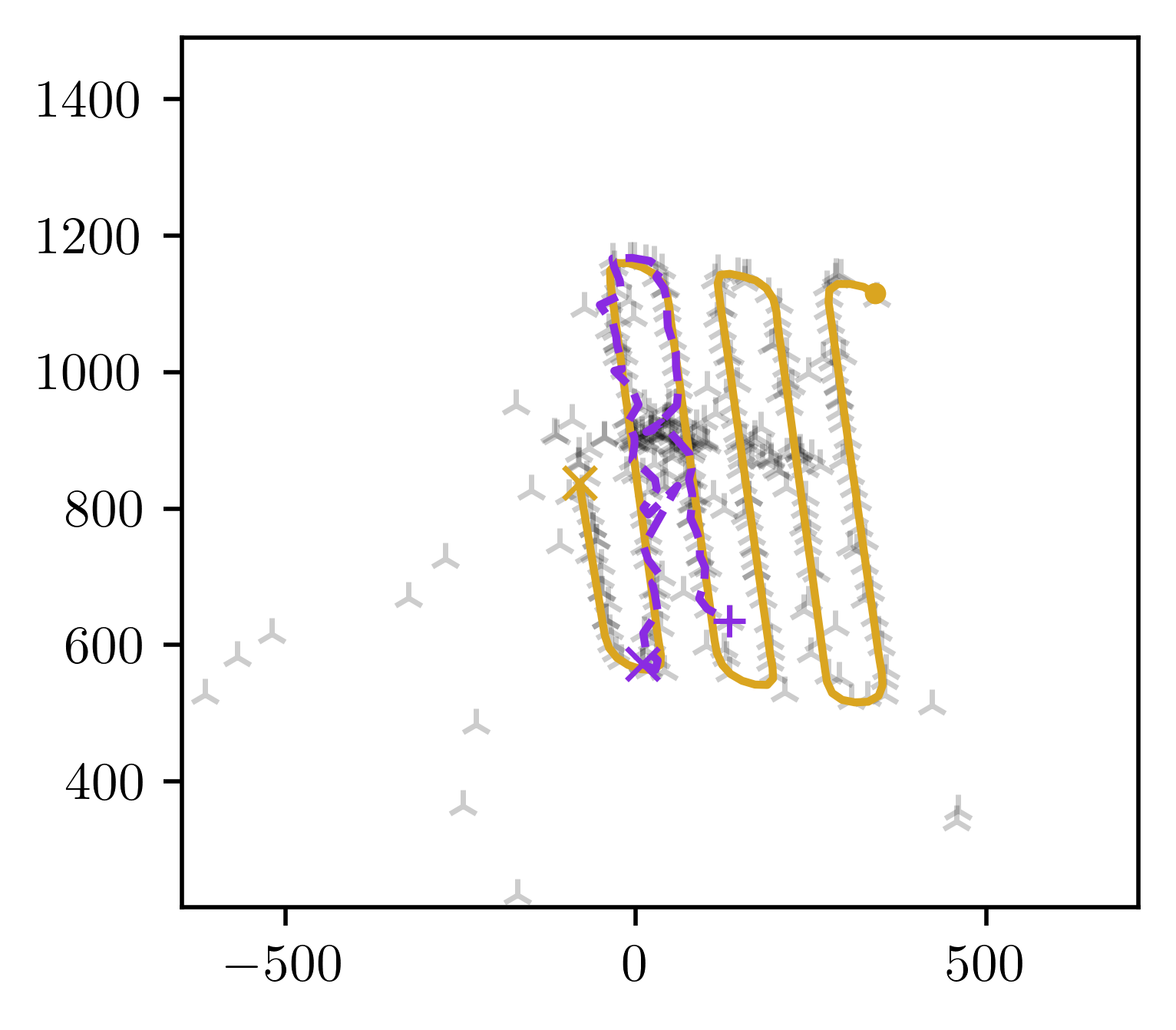}
        \caption{MP-CV}
        \label{subfig:realmpcv}
    \end{subfigure}
    \begin{subfigure}{0.31\linewidth}
        \includegraphics[width=\linewidth]{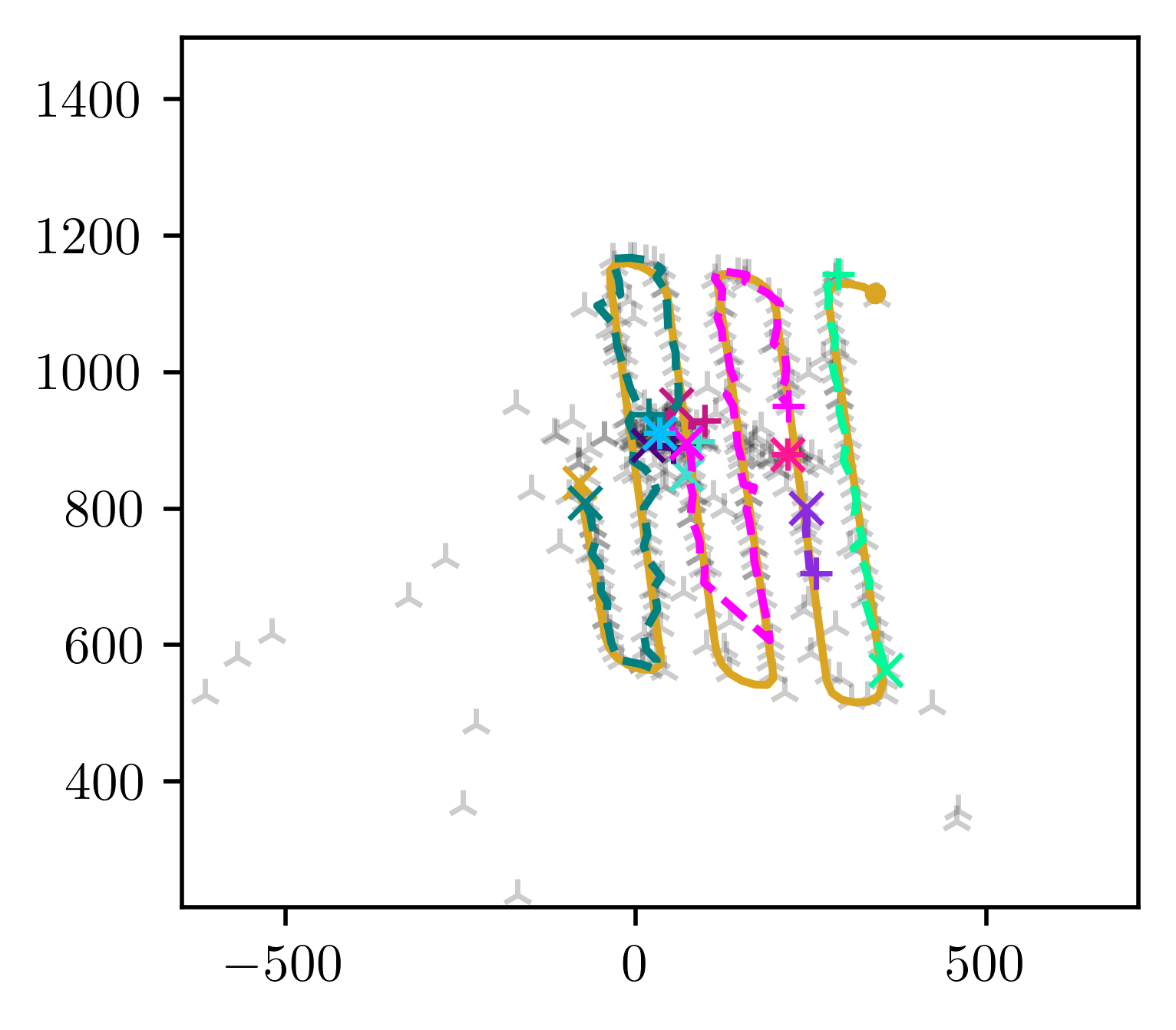}
        \caption{GM-PHD}
        \label{subfig:realgmphd}
    \end{subfigure}
    \caption{Final track estimates from the real data set, with `$\times$' showing a track start, `$\bullet$' a surviving track's end, and `$+$' a deleted track's end. The drone ground truth location is shown in gold and the data are grey stars.}
    \label{fig:realresults9}
\end{figure*}

The results are given in Table \ref{tab:realresults}. The data deviated from the modelling assumptions of all models much more than the synthetic data, and, as such, pose a significantly greater challenge to all methods. The GaPP methods appear to have generalised to the real data better than the MP methods, the GM-PHD and GNN-CV, possibly due to its adaptive nature with respect to the hyperparameters. The reduction in track breaks for GaPP-ReaCtion over GaPP-Class has increased to $\approx30\%$, compared to $\approx15\%$ in the synthetic experiment. The other methods have reduced the gap in this regard, with GaPP-Class having higher m$R$ than the GM-PHD. We see that the GaPPs and DiGiT seems to track the drone better overall, shown by their higher $C$s than for the other methods. It should be acknowledged that DiGiT becomes a much more suitable option in these scenarios, successfully tracking the objects better than all other methods, albeit with more track breaks. That said, the spuriousness was a main drawback of it in the synthetic experiment, but this is not able to be computed here, thus perhaps making DiGiT appear more appealing. However, the fact its \textit{relative} performance significantly improved underlines that the deviation in modelling assumptions of the synthetic data was large, just as it is here for the real data, contrasting that for the other methods.

The outcomes for each model (except GNN-CV) on the exemplar data set are shown in Figure \ref{fig:realresults9}. No method avoids a track break near $(250,850)$, except the MP methods (figs \ref{subfig:realmpimm} and \ref{subfig:realmpcv}), which do little successful tracking. The GaPP methods (figs \ref{subfig:realreaction} and \ref{subfig:realclass}) and DiGiT (fig. \ref{subfig:realdigit}) capture the track well, whilst the GM-PHD (fig. \ref{subfig:realgmphd}) has two further track breaks, and fails to track through the two lowest turns.

\section{Conclusion and Future Work}
\label{sec:conclusion}

This paper introduces two robust and adaptive multi-object trackers, GaPP-Class and GaPP-ReaCtion. Both performed more strongly than chosen state of the art trackers (and in less time) on synthetic data.  They learnt the hyperparameters well, although further development of learning $s^2$ would be beneficial. This included jointly classifying the objects in real time. Overall, the GaPP methods performed best over both synthetic and real data experiments, whilst the MCMC revival kernel of GaPP-ReaCtion helped to reduce track breaks in near-real time by around 30\% on real-world drone surveillance data.

Future work aside from improving the learning of $s^2$ could include a deeper investigation into the robustness of the classification in real scenarios and how to learn the iSE parameters of various object classes. Additionally, this scheme used with a more efficient and sophisticated data clustering procedure could greatly improve the scalability of this algorithm. This could be facilitated by a Sequential MCMC implementation, which is currently under development. Its first iteration is found in the PhD thesis \cite{lydeard2025advanced}, currently under examination.

\appendices

\section{Gaussian Ratios}
\label{app:gaussianratios}

With constant $x$ and $\bar y = \frac1n\sum_{l=1}^ny_l$, the following product of Gaussian ratios is
\cm{
\prod_{l=1}^n\frac{\mathcal N(y_l\mid x,\sigma^2)}{\mathcal N(y_l\mid \bar y,\sigma^2)}
= \exp\left\{-\frac n{2\sigma^2}(\bar y-x)^2\right\}.
}{\label{eq:gaussianratioproduct}}
Multiple such expressions can be combined and simplified, such as in \eqref{eq:finalassoc1.5}. If we have $n_x+n_y=n_z$ as sample sizes, and $n_x\bar x + n_y\bar y = n_z \bar z$ as sample sums, then
\cm{
n_y(\bar z-\bar y)^2 + n_x(\bar z-\bar x)^2
= \frac{n_xn_y}{n_x+n_y}(\bar x-\bar y)^2.
}{\label{eq:simplifydoubleGratioresult}}

\section{Offset-Diagonal Matrices}
\label{app:offsetdiag}

\subsection{Derivations}
\label{subapp:derivations}

Consider matrices $M,N\in\mathbb R^{\alpha\times\alpha}$ of the form
\cm{
M = AI_\alpha + B\bm 1_{\alpha\times\alpha},\! && N = \frac1AI_\alpha - \frac B{\alpha AB + A^2}\bm 1_{\alpha\times\alpha},
}{\label{eq:twoparameterinverse1}}
for some $A\ne0$ and $B\ne -\frac A\alpha$.
Using $\delta_{ij}$ as the Kronecker Delta, $N = M^{-1}$ is shown by computing $MN$ element-wise.
\cm{
(MN)_{ij} 
&= \sum_{k=1}^\alpha (A\delta_{ik}+B)\left(\frac1A\delta_{kj}-\frac B{\alpha AB + A^2}\right)\\
&= \delta_{ij}~~~~\blacksquare
}{\label{eq:twoparameterinverse2}}

The determinant of $M_\alpha := M$ is also computable as
\cm{
\det M_\alpha = A^{\alpha-1}(A+\alpha B),
}{\label{eq:twoparameterdet}}
proven by induction. The cases for $\alpha=1,2$ are trivially computed. The determinant $\det M_{\alpha}$ can be expanded as
\cm{
(A+B)\det M_{\alpha-1} + B\sum_{i=1}^{\alpha-1} (-1)^i \det(M_{\alpha-1} - A\bm e_{i\alpha}\bm e_{i\alpha}^T).
}{\label{eq:twoparameterdetproof1}}
By permuting the `all-B' row $i$ to the top (which is $i-1$ row permutations), one has
\cm{
(-1)^{i-1}\!\det(M_{\alpha-1} \!-\! A\bm e_{i\alpha}\bm e_{i\alpha}^T)\! &=\! \det(M_{\alpha-1} \!-\! A\bm e_{i\alpha}\bm e_{i\alpha}^T)\\
&=\! (\det M_{\alpha-1} \!-\! A\det M_{\alpha-2}).
}{\label{eq:allB}}
Hence, $\det M_{\alpha}$ in \eqref{eq:twoparameterdetproof1} becomes the recurrence relation
\cm{
(A-(\alpha-2)B)\det M_{\alpha-1} + (\alpha-1)AB\det M_{\alpha-2}.
}{\label{eq:twoparameterdetproof2}}
Assuming \eqref{eq:twoparameterdet} for all $\alpha'<\alpha$, substitution into \eqref{eq:twoparameterdetproof2} gives the result
$\forall\,M_\alpha\in\mathbb R^{\alpha\times\alpha}$ with $\alpha\ge 1~\blacksquare$.

\subsection{Simplifications Using Offset-Diagonal Matrices}
\label{subapp:uses}

These allow for simplification of typical Gaussian expressions and manipulations frequently seen in this work. Specifically, a typical linear update step may be required if we have
\cm{
\bm y\mid \bm x&\sim \mathcal N((\bm e_{1d}^T\bm x)\bm 1_n,\Sigma I_n),\\
\bm x &\sim\mathcal N(\check{\bm m},\check V),
}{\label{eq:typicalkalmanupdate1}}
where $d=\dim \bm x, n=\dim \bm y$, and $\bm 1_\alpha$ and $I_\alpha$ mean as they did in the main body of work, above. In such a case, the standard Kalman update equations (e.g., \cite{sarkka2013}) yield
\cm{
&~~~~~~~~~~~~~~~~~~~~~~~~~\bm x\mid\bm y \sim \mathcal N(\bm m,V),\\
&\bm m \!=\! \check{\bm m} \!+\! \check V\bm e_{1d}\bm 1_n^T\!\!\left(\Sigma I_n + (\bm e_{1d}^T\check V\bm e_{1d})\bm 1_{n\times n}\right)^{-1}\!\!(\bm y - \bm 1_n\bm e_{1d}^T\check{\bm m}),\\
&V = \check V - \check V\bm e_{1d}\bm 1_n^T\left(\Sigma I_n + (\bm e_{1d}^T\check V\bm e_{1d})\bm 1_{n\times n}\right)^{-1} \bm 1_n\bm e_{1d}^T\check V.
}{\label{eq:typicalkalmanupdate2}}
Acknowledging that $\check V\bm e_{1d} = \check V_{:,1}$ is the first column of $\check V$, and similar results, \eqref{eq:twoparameterinverse1} ultimately implies
\cm{
\check V\bm e_{1d}\bm 1_n^T\left(\Sigma I_n + (\bm e_{1d}^T\check V\bm e_{1d})\bm 1_{n\times n}\right)^{-1}
= \frac1{n\check V_{1,1}+\Sigma}\check V_{:,1}\bm 1_n^T.
}{\label{eq:typicalkalmanupdate3}}
Then,
\cm{
\bm m = \check{\bm m} + \frac{n(\bar y-\check m_1)}{n\check V_{1,1}+\Sigma}\check V_{:,1},
&& V = \check V - \frac n{n\check V_{1,1}+\Sigma}\check V_{:,1}\check V_{1,:}.
}{\label{eq:typicalkalmanupdate4}}
These update formulae are significantly more efficient than the general ones in \eqref{eq:typicalkalmanupdate2}, scaling in time as $\mathcal O(n)$ rather than $\mathcal O(n^2)$. If  \eqref{eq:typicalkalmanupdate3} is computed directly 
the latter is in fact $\mathcal O(n^3)$.

Likewise, the computation of the marginal likelihood of the above scenario can be simplified. The marginal likelihood is
\cm{
&\mathcal N(\bm y\mid \check m_1\bm 1_n,\check V_{1,1}\bm 1_{n\times n} + \Sigma I_n)\\
&= (2\pi)^{-\frac{n}2}\left(\det \left(\check V_{1,1}\bm 1_{n\times n} + \Sigma I_n\right)\right)^{-\frac12}\\
&~~\times\exp\left\{\!-\frac12 (\bm y \!-\! \check m_1\bm 1_n)^T\!\!\left(\check V_{1,1}\bm 1_{n\times n} \!+\! \Sigma I_n\right)^{-1}\!\!(\bm y \!-\! \check m_1\bm 1_n)\!\right\}.
}{\label{eq:simplifylikelihood1}}
 With $\overline{(y-\bar y)^2} = \frac 1n \sum_{i=1}^n (y_i-\bar y)^2$, the determinant can be directly computed from \eqref{eq:twoparameterdet} and the matrix inverse of \eqref{eq:twoparameterinverse1} can be used to simplify the (negative) exponent of \eqref{eq:simplifylikelihood1} to
\cm{
&\frac12 (\bm y - \check m_1\bm 1_n)^T\!\!\left[\Sigma^{-1}I_n - \frac{\check V_{1,1}}{\Sigma^2 +n\check V_{1,1}\Sigma}\bm 1_{n\times n}\right]\!\!(\bm y - \check m_1\bm 1_n)\\
&~~= \frac n2 \left[\frac1{\Sigma}\overline{(y-\bar y)^2} + \frac1{\Sigma+n\check V_{1,1}}(\bar y - \check m_1)^2\right].
}{\label{eq:simplifylikelihood2}}

\vspace{-5mm}
\section*{Acknowledgment}
This work was funded by the UK Ministry of Defence (MOD) via WSRF under Task WSRF0151. Authors thank the Defence Science and Technology Laboratory (DSTL), UK MOD, for supporting this work. The views and conclusions contained in this paper are of the authors and should not be interpreted as representing the official policies, either expressed or implied, of the UK MOD or the UK Government.

\bibliographystyle{ieeetr}
\bibliography{reaction}
\end{document}